\documentclass[10pt,twocolumn,letterpaper]{article}

\usepackage[pagenumbers]{cvpr}

\usepackage{amsmath,amsfonts,bm}

\def\eqref#1{equation~\ref{#1}}

\def\1{\bm{1}}

\DeclareMathAlphabet{\mathsfit}{\encodingdefault}{\sfdefault}{m}{sl}
\SetMathAlphabet{\mathsfit}{bold}{\encodingdefault}{\sfdefault}{bx}{n}

\usepackage{url}
\usepackage{stfloats} %
\usepackage{lineno}   %
\usepackage[dvipsnames]{xcolor}

\usepackage{comment}
\usepackage{booktabs} 
\usepackage{graphicx}
\usepackage{wrapfig} 
\usepackage{multirow}
\usepackage[most]{tcolorbox}
\usepackage{booktabs}
\usepackage[nosfdefault]{comicneue} %
\usepackage{amssymb}

\usepackage{adjustbox} %

\newcommand{\dcc}[1]{}
\newcommand{\opc}[1]{}
\newcommand{\odc}[1]{}
\newcommand{\gmc}[1]{}
\newcommand{\aec}[1]{}
\newcommand{\esc}[1]{}
\newcommand{\hec}[1]{}
\newcommand{\lmc}[1]{}
\newcommand{\hac}[1]{}

\usepackage{comfortaa} %
\usepackage{tikz}
\usetikzlibrary{arrows.meta,positioning,calc,fit,backgrounds,shapes.geometric,
                decorations.pathmorphing,decorations.pathreplacing,svg.path}

\makeatletter\providecommand{\setcaptypefigure}{\def\@captype{figure}}\makeatother %

\definecolor{cvprblue}{rgb}{0.21,0.49,0.74}
\usepackage[pagebackref,breaklinks,colorlinks,allcolors=cvprblue]{hyperref}
\hypersetup{pdftitle={Open-Form Geometric Style Transfer},
  pdfauthor={Adar Elad, Gal Metzer, Omer Dahary, Min Lu, Daniel Cohen-Or}}

\title{Open-Form Geometric Style Transfer}

\author{%
Adar Elad$^{1*}$ \hspace{1.5em} Gal Metzer$^{1}$ \hspace{1.5em} Omer Dahary$^{1}$ \hspace{1.5em}
Min Lu$^{2}$ \hspace{1.5em} Daniel Cohen-Or$^{1}$\\[4pt]
$^{1}$Tel Aviv University \hspace{3em} $^{2}$Shenzhen University\\[4pt]
{\tt\small \{elad.research, gal.metzer, omer11a, lumin.viz, cohenor\}@gmail.com}
}

\begin{document}

\twocolumn[{%
\renewcommand\twocolumn[1][]{#1}%
\maketitle
\begin{center}
\setcaptypefigure
\input{figures/teaser}
\end{center}
\vspace{1em}
}]
{\renewcommand{\thefootnote}{}\footnotetext{$^{*}$Corresponding author}}

\begin{abstract}
A geometric style learned from one family of shapes is typically applied to
new shapes from the same family. We consider a more challenging setting:
learning the geometric design language of a stylized Latin alphabet and
applying it to new forms, from Greek glyphs to entirely different structures
such as human poses. We introduce Open-Form Geometric Style Transfer,
where the forms to which a learned geometric style can be applied are not
restricted to the shape family in which that style was observed.
This setting is deceptively challenging: the difficulty is not merely learning
a style, but learning the style without learning the shape family in which it
was observed. We represent target forms by structural sketches and disentangle
structural adherence from style realization using a Geometry LoRA and a
style-specific LoRA. Critically, open-form transfer requires the Geometry LoRA
to be universal. To achieve this, we devise a geometry-enrichment
strategy that constructs a large and purposefully diverse collection of
synthetic structural sketches spanning a broad range of unfamiliar geometries.
Our experiments and ablations demonstrate the importance of both
disentanglement and this universal geometry prior, enabling geometric styles
learned from glyphs, pictograms, and other shape families to be transferred to
substantially different, unseen forms.
\end{abstract}

\section{Introduction}
\input{figures/open_form_examples}

Style transfer is a long-standing and intriguing problem in visual computing:
given examples that contain both content and style, the goal is to capture the
essence of the style, separate it from the particular content on which it is
observed, and apply it to new content. The separation of style from content has
been studied for decades~\citep{efros1999texture,efros2001quilting}, and has more
recently become central to neural style transfer and generative
modeling~\citep{gatys2015neural,voynov2023sketch,stylealigned}.
At the heart of the problem lies the ability to distinguish \emph{what is
represented} from \emph{how it is expressed}.

In this work, we focus on \emph{geometric style}, where style governs how an
underlying structure is geometrically realized. A canonical example is a font:
different glyphs have different structures, yet their geometric realizations
follow a coherent design language that makes them immediately recognizable as
belonging to the same font. More generally, such geometric styles cannot be
inferred from a single example; it is revealed through a collection of different
shapes that share the same design language. Early work has considered related
notions of separating style from content and transferring geometric style across
shapes~\citep{tenenbaum2000separating,freeman2003}.

Existing approaches \citep{park2021mxfont,fontdiffuser,vecfusion} typically learn and apply such styles within a shared domain: a style observed on letters is transferred to other letters, and a style observed on human figures is applied to other human figures. 
We remove this restriction and consider transferring a geometric style to an \emph{open space of forms}. 
As illustrated in Figure~\ref{fig:teaser}, a style learned from a
Latin alphabet may be applied to Greek glyphs, or more radically, to human
poses; conversely, a geometric style learned from pictograms may be applied to
glyphs. We call this new setting \emph{Open-Form Geometric Style Transfer},
where the forms to which a learned geometric style can be applied are not
restricted to the shape family in which that style was observed. We focus on
forms whose underlying structure can be represented by a sparse structural
sketch, which serves as the geometry to be styled.

The open-form setting, illustrated in Figure~\ref{fig:open}, is deceptively challenging.
The difficulty is not simply learning a style, but \emph{learning the style
without learning the shape family in which that style was observed}. 
Existing font-generation methods can transfer styles to unseen
glyphs~\cite{fontdiffuser,vecfusion,skeletonfont,fluxfont}, and some explicitly
separate character content from style. However, their structural component is
learned and evaluated largely within the glyph domain, often within a single
writing system. This distinction becomes critical when the target form leaves
the source domain.
A straightforward LoRA trained on a limited styled shape family learns an entangled
style with the geometries observed at training (Sec.~\ref{sec:geometric-generalization}), causing it
to drift toward familiar source-domain shapes when presented with substantially
different target geometries.

Our approach addresses this entanglement by separating \emph{structural adherence from style realization}.
A Geometry LoRA is responsible for faithfully following the target structural sketch serving as a geometric-adherence prior, while a Style LoRA learns how that geometry should be realized in the desired style. 
Accordingly, the Geometry LoRA must be \textbf{universal}: generalize across the vast universe of styles, responding to structure rather than shape identity or residual stylistic cues.
Acquiring such a universal geometry prior is itself non-trivial: a corpus of ordinary glyphs provides a limited and familiar geometry vocabulary, allowing the model to recognize known shapes rather than truly follow the input structure. 
We therefore introduce a geometry-enrichment strategy (GEE Sec.~\ref{sec:gee}) that systematically leverages SVG font data to construct a large collection of unfamiliar synthetic geometries, untangled from their source style. 
This strategy proves vital for successfully learning a universal geometry LoRA.

Our contributions are threefold:
\begin{itemize}
  \item We introduce \textbf{Open-Form Geometric Stylization}, where a geometric style learned from one shape domain is applied to universally arbitrary forms. 
  \item We show that open-form stylization requires separating structural adherence from style realization. 
  \item Finally, we introduce a geometry-enrichment strategy for constructing the broad and diverse structural training distribution needed to acquire this universal prior. 
\end{itemize}
We evaluate these ideas through controlled ablations and increasingly challenging transfers, from held-out glyphs and cross-alphabet transfer to unrelated forms such as human poses, demonstrating that a geometric style can be extended well beyond the family of shapes on which it was observed.

\section{Background and Related Work}
\label{sec:related}
Our work builds on three lines of research: the separation of style and
content, font and glyph style transfer, and structurally conditioned
generation. Below we review each line, clarify how our notion of
\emph{geometric} style relates to them, and how they relate to our approach.

\subsection{Style and Content}

Our work is closely connected to the seminal work of Tenenbaum and
Freeman~\cite{tenenbaum2000separating} on separating style and content.
Using bilinear models, they showed that a common style can be inferred from
multiple observations and applied to new content. Particularly relevant to
our setting is their treatment of fonts, where style is shared across
different letters and can be used to synthesize previously unseen characters.
This provides an early and elegant formulation of style as a factor that can
be learned across a family of shapes and separated from their individual
geometry. Our work builds on this perspective, but considers an open-form
setting in which the new content need not belong to the same family: a
geometric style learned from one alphabet may be applied to another alphabet,
or even to a different class of structures such as human poses.

More broadly, style transfer has a long history in computer vision and
graphics, with much of the literature focusing on style as visual appearance.
Early example-based methods~\citep{efros1999texture,efros2001quilting} synthesized and transferred texture by sampling local neighborhoods or patches
from example images.
\cite{gatys2015neural} later introduced neural style transfer, demonstrating that content and artistic style can be represented separately within a deep network and recombined to render new content in the appearance of a reference image. Recent diffusion-based methods
such as StyleAligned~\cite{stylealigned}, B-LoRA~\cite{blora},
InstantStyle~\citep{instantstyle}, and ZipLoRA~\citep{shah2023ZipLoRA}
substantially extend the quality and flexibility of reference-based style
transfer. Across these works, style is predominantly expressed through visual
appearance, including texture, color, material, brushwork, and shading.

Our notion of \emph{geometric style} is different. Here, style determines how
an underlying structure is geometrically realized: how curves are formed,
junctions are connected, strokes vary in thickness, gaps are introduced, and
structural elements terminate or interact. Such a style is naturally revealed
by a \emph{set} of related shapes rather than by a single reference image.
Glyphs provide a particularly clean instance of this setting: individual
characters differ in structure, while together they reveal a consistent
geometric design language. Related notions of transferring geometric style
across shapes were also explored in early work by~\cite{freeman2003}.
More recently, geometric stylization has also been explored for 3D shapes ~\cite{choi2026imageguidedgeometricstylization3d}.

Closely related, \emph{Abstraction in Style} (AiS)~\citep{lu2026abstraction} also
separates geometry from appearance using two specialized LoRA adapters.
However, AiS learns style-specific geometric abstraction, whereas our
Geometry LoRA is universal, enabling a learned geometric style to generalize
to unseen structural sketches across shape families.

\subsection{Font and Glyph Style Transfer}
Modern font-generation methods provide the closest setting to our work, as
they explicitly transfer a common style across glyphs with different
structures. FontDiffuser~\cite{fontdiffuser} formulates one-shot
diffusion-based font generation, using multi-scale content aggregation to
preserve character structure and contrastive learning to capture style. It
generalizes to unseen characters and also demonstrates cross-lingual
generation. MX-Font~\cite{park2021mxfont} addresses few-shot font generation using multiple
localized experts that disentangle content and style while capturing different local style characteristics. 
It also demonstrates cross-lingual generalization to unseen writing systems.
VecFusion~\cite{vecfusion} approaches font generation in the
vector domain, explicitly modeling glyph topology, paths, and B\'ezier control
points, and supports applications including glyph completion and few-shot
style transfer. Skeleton-Guided Diffusion~\cite{skeletonfont} is particularly
related to our structural representation, using character skeletons to guide
diffusion and preserve structure during font generation. StrokeStyles~\cite{berio2022strokestyles} decomposes font glyphs into
meaningful stroke structures and uses this representation for consistent
geometric stylization across characters.

FLUX-Font~\cite{fluxfont} is particularly close to our work architecturally.
It adapts a pre-trained diffusion transformer using separate LoRA
modules~\cite{hu2022lora} for glyph content and style, explicitly separating
the two factors rather than learning them through a single adapter. Our method
shares this general motivation, but separates structural adherence from style
realization hierarchically: a universal Geometry LoRA is first trained to
follow arbitrary structural sketches, and each Style LoRA is subsequently
learned with this geometry prior active and fixed.

Despite their differences, these methods share an important restriction:
their structural representations and learned priors are designed for the
glyph domain. Their goal is to generalize from observed characters to other
characters, possibly across writing systems, but the target remains a glyph.
Open-form transfer removes this restriction. The target structural sketch may
describe a character, a pictogram, a human pose, or another unrelated form.
Consequently, separating content from style within the font domain is not
sufficient; the geometry-following component must itself generalize beyond
the shape family in which the style was observed.

\definecolor{geeGreen}{RGB}{55,125,95}
\definecolor{geoBlue}{RGB}{65,135,175}
\definecolor{styleOrange}{RGB}{206,114,80}
\definecolor{tileViolet}{RGB}{120,80,165}
\definecolor{tileTeal}{RGB}{30,130,150}
\definecolor{inferInk}{RGB}{91,83,130}

\tikzset{
  panel/.style={rounded corners=5pt, line width=.4pt},
  tile/.style ={rounded corners=2.5pt, line width=.35pt, fill=white},
  box/.style  ={rounded corners=3pt, line width=.45pt},
  adapter/.style={box,rounded corners=1.8pt},
  flow/.style ={-{Straight Barb[length=3pt,width=3pt]}, line width=.45pt, black!85},
  lbl/.style  ={inner sep=0pt, outer sep=0pt},
}

\providecommand{\snow}{\tikz[baseline=-.5ex,scale=.075]{%
  \foreach \a in {0,60,...,300}{%
    \draw[line width=.4pt,cyan!60!blue,line cap=round](0,0)--(\a:1);
    \draw[line width=.4pt,cyan!60!blue,line cap=round](\a:.55)--++(\a+40:.28);
    \draw[line width=.4pt,cyan!60!blue,line cap=round](\a:.55)--++(\a-40:.28);}}}
\definecolor{flameTop}{RGB}{250,131,15}
\definecolor{flameBottom}{RGB}{189,5,6}
\newcommand{\flameicon}[1]{%
  \tikz[baseline=0pt]{
    \begin{scope}[x=1pt,y=1pt,cm={{#1/180.48*0.024},0,0,{#1/180.48*0.024},(0pt,0pt)}]
      \shade[top color=flameTop,bottom color=flameBottom] svg{M 1821 7443 c 8 -10 32 -40 54 -68 121 -149 197 -298 227 -448 57 -284 0 -594 -190 -1042 -53 -125 -277 -575 -310 -623 -10 -15 -47 -79 -82 -142 -34 -63 -68 -122 -75 -130 -7 -8 -27 -44 -45 -80 -19 -36 -43 -78 -55 -95 -19 -26 -49 -76 -89 -150 -6 -11 -51 -86 -101 -168 -49 -81 -100 -164 -111 -185 -12 -20 -60 -100 -107 -177 -143 -234 -445 -740 -509 -853 -10 -18 -26 -44 -34 -58 -9 -13 -13 -24 -8 -24 4 0 0 -6 -9 -12 -35 -27 -216 -425 -203 -446 3 -6 0 -12 -8 -15 -7 -3 -16 -18 -19 -34 -3 -15 -15 -63 -26 -106 -11 -43 -17 -85 -14 -93 3 -8 1 -14 -5 -14 -11 0 -18 -38 -34 -206 -8 -90 -9 -149 0 -235 15 -152 23 -199 32 -199 4 0 8 -15 7 -32 -1 -40 43 -189 53 -183 5 3 6 -5 2 -17 -4 -16 -2 -19 8 -13 8 5 11 4 6 -3 -8 -14 29 -108 42 -105 6 1 8 -6 5 -17 -3 -12 1 -22 11 -26 9 -3 13 -10 11 -15 -8 -12 81 -165 94 -162 6 2 8 -3 6 -10 -7 -18 173 -252 193 -252 10 0 16 -7 14 -17 -2 -15 13 -32 31 -33 4 0 7 -4 7 -10 1 -17 63 -75 73 -68 6 3 7 2 4 -4 -3 -5 9 -25 28 -43 19 -19 35 -30 35 -25 0 5 7 -1 16 -13 8 -12 20 -19 26 -16 6 4 8 3 5 -3 -8 -13 112 -115 126 -106 6 3 7 1 3 -5 -8 -13 28 -35 39 -24 5 4 5 2 2 -5 -7 -11 101 -95 238 -184 86 -56 115 -72 115 -62 0 4 6 -1 13 -10 25 -33 331 -192 349 -181 6 4 8 3 5 -3 -8 -13 20 -26 35 -16 6 3 8 2 5 -4 -4 -6 13 -16 36 -23 23 -8 47 -17 52 -21 6 -4 19 -9 30 -11 11 -1 23 -7 26 -11 4 -5 18 -10 32 -10 14 -1 23 -5 21 -8 -6 -10 45 -21 58 -13 7 4 8 3 4 -4 -8 -14 98 -47 115 -36 8 4 9 3 5 -4 -4 -7 -1 -12 7 -12 8 0 14 10 13 24 -1 30 -253 329 -269 319 -6 -3 -7 -1 -3 6 9 14 -12 43 -25 35 -5 -3 -8 -2 -7 3 5 17 -33 74 -45 67 -7 -4 -9 -4 -5 1 11 11 -11 47 -24 39 -6 -4 -8 -3 -5 3 4 6 -20 51 -53 101 -33 49 -90 150 -126 223 -37 74 -70 135 -74 137 -4 2 -5 12 -2 24 4 16 2 19 -8 13 -9 -5 -11 -4 -6 4 4 6 1 33 -8 59 -24 71 -59 194 -61 217 -36 310 -28 477 35 700 13 44 23 88 24 98 0 9 4 17 9 17 5 0 6 7 3 16 -3 8 -2 12 4 9 6 -3 10 0 10 7 0 21 59 170 122 306 21 46 38 89 38 96 0 7 4 16 9 21 22 23 101 187 95 197 -4 6 -3 8 3 5 11 -7 76 105 68 117 -3 3 -2 4 2 1 7 -6 106 154 106 172 0 7 3 11 5 8 3 -2 46 57 96 133 112 167 175 259 218 316 18 24 40 54 48 66 8 12 47 67 87 121 105 143 310 447 385 569 60 99 179 328 205 395 6 17 23 60 37 97 14 36 37 108 51 160 33 122 34 287 4 439 -19 99 -82 298 -120 384 -9 22 -36 83 -59 135 -73 165 -231 453 -344 630 -55 84 -230 347 -261 389 -11 16 -44 62 -72 102 -48 68 -130 174 -232 304 -54 68 -136 155 -145 155 -5 0 -2 -8 5 -17z m -1111 -6603 c 0 -5 -2 -10 -4 -10 -3 0 -8 5 -11 10 -3 6 -1 10 4 10 6 0 11 -4 11 -10z M 3430 5024 c 0 -6 11 -36 24 -65 46 -104 57 -173 58 -339 0 -220 -19 -312 -118 -590 -40 -114 -174 -372 -269 -520 -44 -68 -52 -81 -125 -190 -46 -70 -249 -358 -352 -500 -32 -44 -58 -81 -58 -84 0 -2 -28 -46 -63 -98 -71 -103 -118 -183 -114 -189 1 -2 0 -3 -2 -2 -7 4 -70 -101 -65 -109 3 -5 1 -8 -4 -8 -22 0 -208 -333 -196 -352 3 -4 1 -8 -5 -8 -6 0 -7 -7 -4 -17 4 -11 3 -14 -5 -9 -7 4 -15 -9 -22 -36 -6 -24 -16 -56 -21 -73 -17 -49 -23 -249 -10 -328 18 -108 34 -167 43 -161 4 3 5 -2 2 -10 -4 -9 -1 -16 6 -16 7 0 10 -6 7 -14 -8 -20 23 -97 36 -89 6 3 7 1 3 -5 -9 -15 15 -63 27 -55 6 4 8 0 4 -9 -7 -19 63 -136 126 -212 84 -102 89 -107 111 -100 14 3 17 2 8 -4 -10 -7 -10 -11 2 -21 11 -9 16 -9 22 0 5 8 3 9 -6 4 -10 -6 -12 -4 -7 8 8 21 -10 70 -22 63 -5 -3 -8 0 -8 7 1 6 -5 44 -12 83 -13 69 -12 117 5 264 4 30 10 57 14 60 4 3 15 30 24 60 25 78 66 176 99 238 16 29 26 60 23 68 -3 8 -2 13 2 10 12 -7 64 92 56 106 -4 6 -3 8 3 5 13 -8 67 71 65 96 0 10 4 15 10 11 7 -5 9 -1 5 9 -4 10 -2 16 4 14 13 -3 150 189 146 205 -1 7 3 12 10 13 17 0 55 53 47 66 -4 7 -3 9 3 6 13 -8 65 57 55 68 -4 3 -2 5 4 3 15 -5 77 70 79 95 1 12 6 22 11 22 27 0 290 358 388 528 36 62 166 323 166 332 0 4 7 21 15 38 24 53 65 168 86 242 11 39 28 95 37 125 22 73 24 471 3 570 -13 59 -56 198 -110 354 -42 119 -241 483 -241 440z m -859 -3387 c -10 -9 -11 -8 -5 6 3 10 9 15 12 12 3 -3 0 -11 -7 -18z M 4290 4156 c 0 -5 13 -36 28 -70 69 -146 98 -357 81 -579 -5 -73 -13 -145 -18 -162 -6 -16 -13 -50 -16 -75 -22 -139 -99 -392 -170 -555 -107 -243 -182 -361 -403 -630 -50 -60 -88 -114 -85 -120 4 -5 3 -7 -2 -3 -11 11 -36 -12 -28 -26 3 -6 3 -8 -2 -4 -9 9 -11 7 -120 -118 -105 -120 -127 -149 -117 -159 4 -5 2 -5 -5 -1 -13 7 -45 -32 -37 -46 3 -4 -1 -8 -8 -8 -16 0 -58 -60 -58 -84 0 -9 -4 -15 -9 -11 -11 6 -51 -47 -56 -73 -1 -9 -7 -16 -13 -15 -15 3 -74 -96 -66 -109 5 -7 2 -8 -6 -3 -10 6 -12 4 -7 -8 3 -8 0 -18 -7 -20 -14 -5 -49 -74 -49 -94 0 -7 -4 -13 -8 -13 -8 0 -13 -11 -44 -99 -13 -35 -21 -71 -18 -80 4 -9 2 -13 -2 -8 -17 15 -73 -238 -96 -428 -9 -85 -11 -261 -3 -304 4 -17 11 -28 17 -24 5 3 8 0 5 -8 -8 -23 15 -39 30 -20 7 9 10 10 6 2 -5 -9 -3 -12 4 -7 19 12 64 95 57 106 -3 5 -1 10 5 10 6 0 8 5 4 12 -4 7 -3 8 4 4 13 -8 65 87 54 99 -4 5 -1 4 7 0 11 -6 13 -4 8 9 -4 9 -3 15 2 11 11 -6 53 49 44 59 -3 3 -1 6 6 6 7 0 9 5 5 12 -4 6 -3 8 3 5 12 -8 120 120 115 135 -1 5 1 7 6 4 12 -7 64 48 55 59 -5 4 -3 5 3 2 6 -3 47 31 91 76 44 45 105 104 134 132 30 27 59 58 66 70 7 11 12 16 13 11 0 -5 24 13 52 40 29 27 51 54 50 59 -2 6 0 7 4 3 10 -10 155 122 147 135 -3 6 -1 7 5 3 6 -4 55 37 111 93 93 92 120 124 89 105 -9 -5 -9 -2 1 9 15 18 26 23 16 7 -3 -5 -3 -10 2 -9 8 0 102 107 186 211 26 33 44 65 40 71 -3 7 -3 9 2 5 10 -10 78 79 70 92 -4 6 -1 9 5 8 7 -2 19 11 29 27 9 17 35 62 57 100 86 148 127 248 175 425 47 170 62 315 56 540 -9 376 -62 529 -329 950 -22 36 -46 74 -52 85 -6 11 -19 31 -27 45 -9 14 -33 52 -53 85 -20 33 -36 55 -36 51z m -440 -2990 c 0 -3 -4 -8 -10 -11 -5 -3 -10 -1 -10 4 0 6 5 11 10 11 6 0 10 -2 10 -4z};
    \end{scope}}}

\providecommand{\flame}{\hspace{1pt}\raisebox{-1.1pt}{\flameicon{6.6}}}

\providecommand{\pairimg}{0.88cm}%
\providecommand{\soloimg}{1.38cm}%

\providecommand{\pairtile}[5]{%
  \begin{scope}[shift={(#1,#2)}]
    \draw[tile,draw=#3!50] (0,0) rectangle (2.04,1.02);
    \draw[black!35,line width=.25pt] (1.02,0)--(1.02,1.02);
    \node[lbl] at (0.51,0.51) {\includegraphics[width=\pairimg]{figures/assets/method_overview/#4}};
    \node[lbl] at (1.53,0.51) {\includegraphics[width=\pairimg]{figures/assets/method_overview/#5}};
  \end{scope}}
\providecommand{\pairtilebig}[5]{%
  \begin{scope}[shift={(#1,#2)}]
    \draw[tile,draw=geeGreen!60] (0,0) rectangle (1.55,0.75);
    \draw[black!35,line width=.25pt] (0.775,0)--(0.775,0.75);
    \node[lbl] at (0.3875,0.375) {\includegraphics[width=0.66cm]{figures/assets/method_overview/#4}};
    \node[lbl] at (1.1625,0.375) {\includegraphics[width=0.66cm]{figures/assets/method_overview/#5}};
  \end{scope}}
\providecommand{\sqtile}[4]{%
  \begin{scope}[shift={(#1,#2)}]
    \node[lbl] at (0.75,0.75) {\includegraphics[width=\soloimg]{figures/assets/method_overview/#4}};
  \end{scope}}
\providecommand{\ditstack}[3]{%
  \begin{scope}[shift={(#1,#2)}]
    \draw[box,draw=inferInk!45,fill=inferInk!4] (0,0) rectangle (1.70,1.65);
    \node[lbl,font=\comfortaa\tiny] at (.85,1.37) {DiT \snow};
    \draw[adapter,draw=geoBlue!75,fill=geoBlue!16] (.14,.73) rectangle (1.56,1.07);
    \node[lbl,font=\comfortaa\tiny] at (.85,.90) {UGL \snow};
    \draw[adapter,draw=styleOrange!75,fill=styleOrange!18] (.14,.20) rectangle (1.56,.58);
    \node[lbl,font=\comfortaa\tiny] at (.85,.39) {Style LoRA #3};
  \end{scope}}

\providecommand{\dial}[6]{%
  \begin{scope}[shift={(#1,#2)}]
    \draw[line width=1.6pt,black!15,line cap=round] (0,0)--(#3,0);
    \draw[line width=1.6pt,#5,line cap=round] (0,0)--({#3*#4},0);
    \fill[#5] ({#3*#4},0) circle (.055);
    \node[anchor=south,inner sep=1.5pt] at ({#3/2},.05) {\tiny #6};
  \end{scope}}

\begin{figure*}[!tp]
\centering
\begin{tikzpicture}[x=1.25cm,y=1cm,font=\comfortaa]

\draw[panel,draw=black!45,fill=white] (0,0.15) rectangle (13.90,1.75);
\node[anchor=west,lbl] at (0.16,1.52)
  {\small\upshape (i) Geometry Enrichment Engine};

\node[lbl] at (0.97,1.12) {\tiny };
\draw[box,draw=black!25,fill=white] (0.12,0.27) rectangle (1.82,1.02);
\foreach \i/\x in {1/0.40, 2/0.78, 3/1.16, 4/1.54}
  {\node[lbl] at (\x,0.645) {\includegraphics[width=0.34cm]{figures/assets/method_overview/vocab_\i}};}

\draw[flow] (1.82,0.645)--(2.00,0.645);

\draw[box,draw=geeGreen!65,fill=geeGreen!12] (2.04,0.27) rectangle (2.94,1.02);
\node[lbl] at (2.49,0.645) {\scriptsize\upshape\color{geeGreen!70!black} GEE};

\draw[flow] (2.94,0.645)--(3.26,0.645);

\node[anchor=west,lbl] at (3.30,1.22) {\tiny\upshape enriched pairs};
\pairtilebig{3.30}{0.27}{geoBlue}{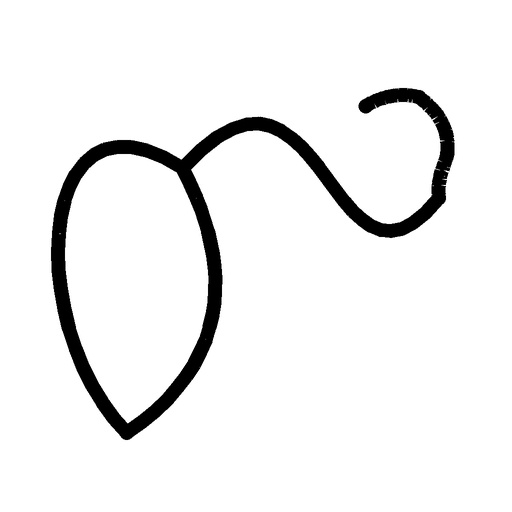}{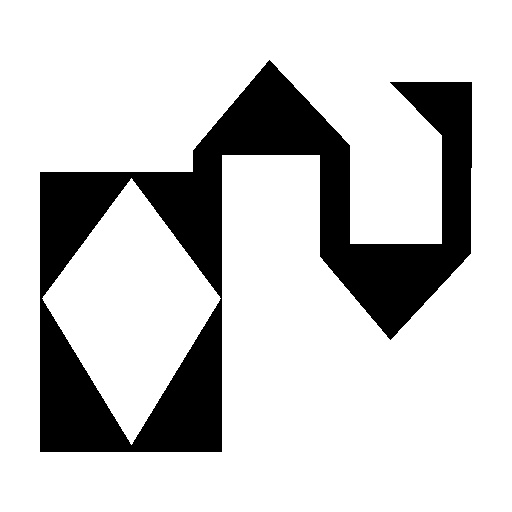}
\pairtilebig{4.99}{0.27}{geeGreen}{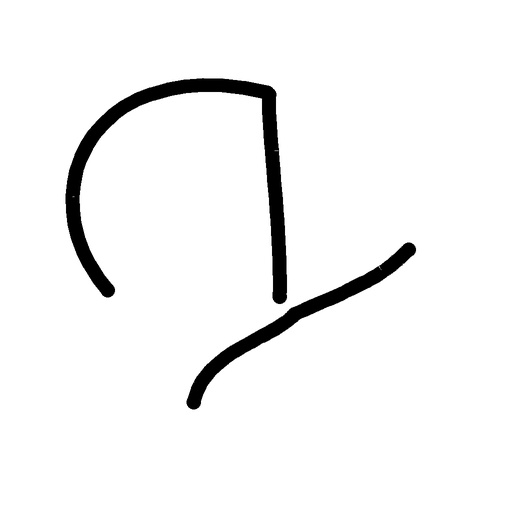}{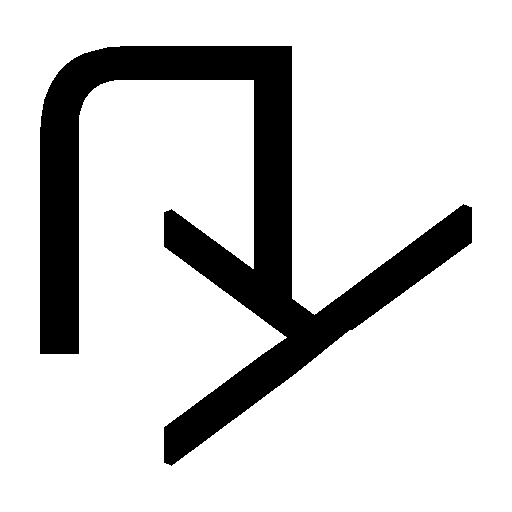}
\pairtilebig{6.68}{0.27}{tileViolet}{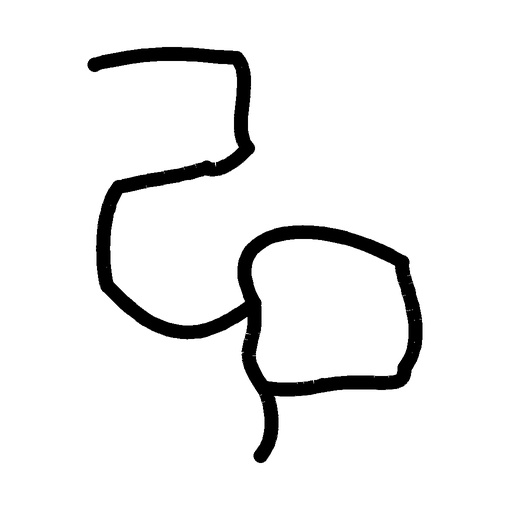}{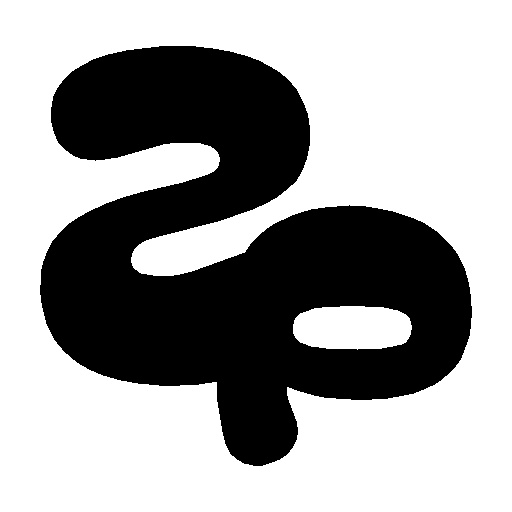}
\pairtilebig{8.37}{0.27}{tileTeal}{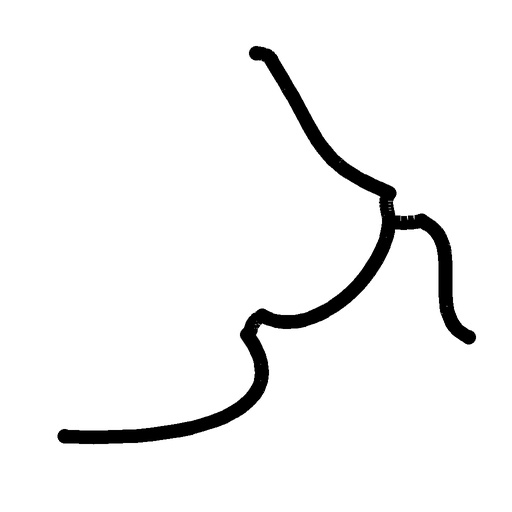}{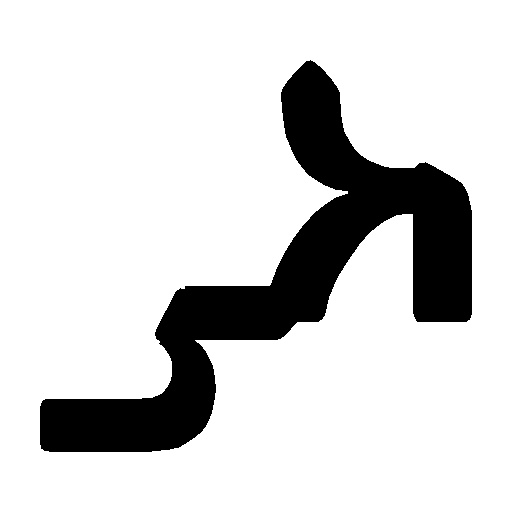}
\pairtilebig{10.06}{0.27}{tileViolet}{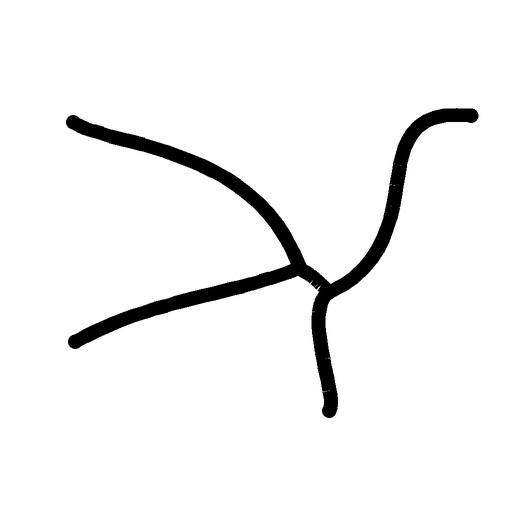}{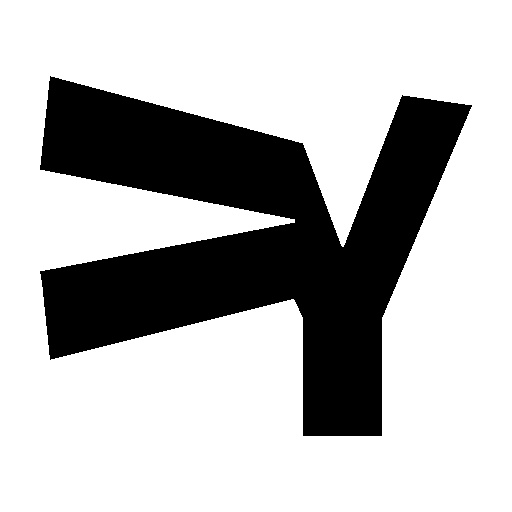}
\pairtilebig{11.75}{0.27}{geoBlue}{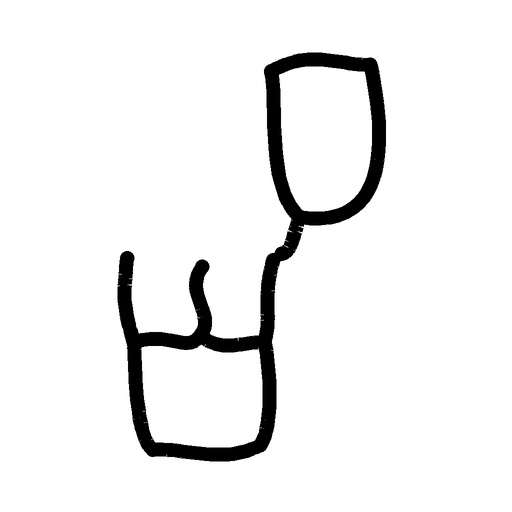}{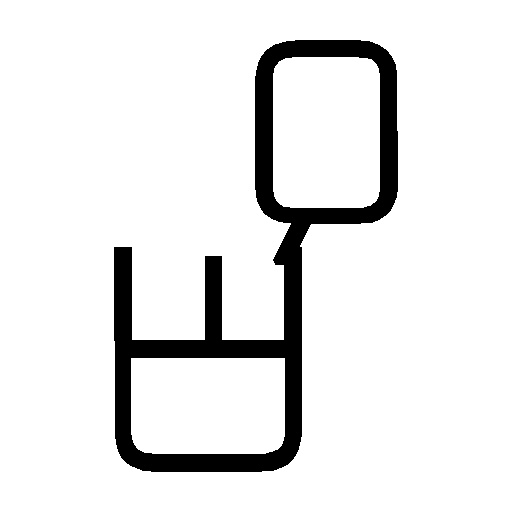}
\node[lbl] at (13.60,0.645) {\tiny$\cdots$};

\node[anchor=west,lbl] at (7.28,1.97)
  {\tiny\upshape geometry reinforcement};

\draw[panel,draw=black!45,fill=white] (0,2.19) rectangle (2.70,5.65);
\node[lbl,text=black,font=\comfortaa\small\upshape] at (1.35,5.34) {(ii) UGL Training};
\node[lbl,text=black!60,font=\comfortaa\tiny] at (1.35,5.02) {Section~\ref{sec:ugl}};

\draw[box,draw=inferInk!45,fill=inferInk!4] (0.18,2.97) rectangle (2.52,4.62);
\node[lbl,font=\comfortaa\tiny] at (1.35,4.34) {DiT \snow};
\draw[adapter,draw=geoBlue!75,fill=geoBlue!16] (0.34,3.17) rectangle (2.36,4.04);
\node[align=center,lbl,font=\comfortaa\fontsize{6}{7}\selectfont] at (1.35,3.60)
  {Universal\\Geometry\\LoRA \flame};

\draw[panel,draw=black!45,fill=white] (2.90,2.19) rectangle (8.30,5.65);
\node[lbl,text=black,font=\comfortaa\small\upshape] at (5.60,5.34)
  {(iii) Style LoRA Training};
\node[lbl,text=black!60,font=\comfortaa\tiny] at (5.60,5.02)
  {Section~\ref{sec:style-lora}};

\node[lbl,font=\comfortaa\tiny\upshape] at (4.22,4.72) {target style examples};
\draw[tile,draw=styleOrange!42] (3.28,2.34) rectangle (5.32,4.46);
\draw[tile,draw=styleOrange!55] (3.24,2.38) rectangle (5.28,4.50);
\pairtile{3.20}{3.52}{styleOrange}{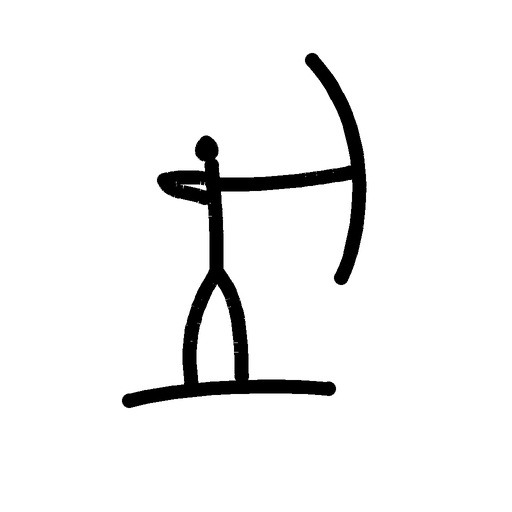}{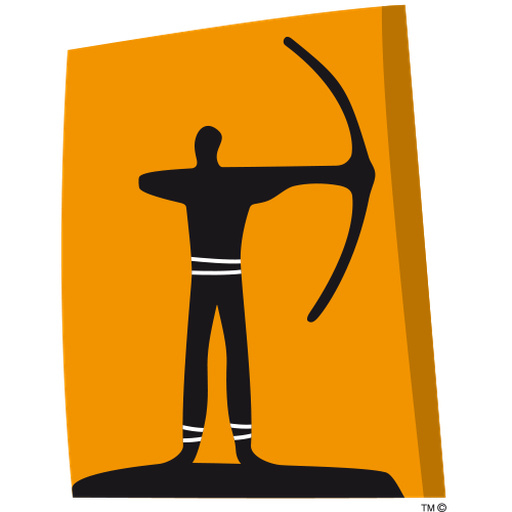}
\pairtile{3.20}{2.42}{styleOrange}{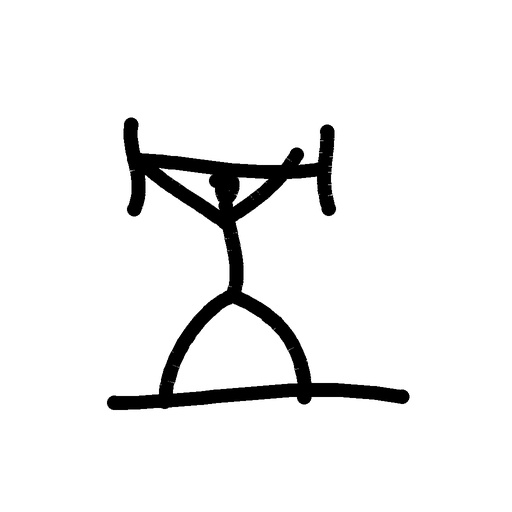}{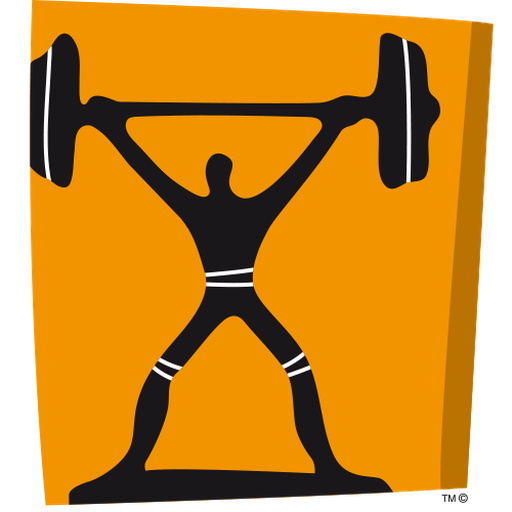}

\draw[black!85,line width=.45pt,line cap=round]
  (5.36,4.03) .. controls (5.69,4.03) and (5.69,3.48) .. (6.02,3.48);
\draw[black!85,line width=.45pt,line cap=round]
  (5.36,2.93) .. controls (5.69,2.93) and (5.69,3.48) .. (6.02,3.48);
\draw[flow] (6.02,3.48) -- (6.29,3.48);
\ditstack{6.30}{2.655}{\flame}

\draw[flow] (2.70,3.80)--(2.90,3.80);

\draw[panel,draw=black!45,fill=white] (8.50,2.19) rectangle (13.90,5.65);
\node[lbl,text=black,font=\comfortaa\small\upshape] at (11.20,5.34) {(iv) Inference};
\node[lbl,text=black!60,font=\comfortaa\tiny] at (11.20,5.02) {Section~\ref{sec:inference}};

\node[align=center,lbl,font=\comfortaa\tiny\upshape] at (9.30,4.47) {unseen sketch};
\sqtile{8.55}{2.73}{inferInk!35}{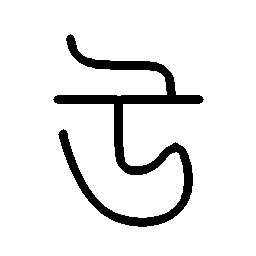}
\draw[flow] (10.03,3.48)--(10.33,3.48);

\ditstack{10.35}{2.655}{\snow}

\draw[flow] (12.07,3.48)--(12.37,3.48);
\node[lbl,font=\comfortaa\tiny\upshape] at (13.10,4.47) {styled output};
\sqtile{12.35}{2.73}{styleOrange!55}{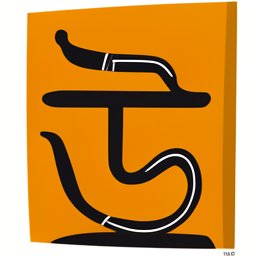}

\draw[flow,rounded corners=2.5pt] (4.20,1.75) -- (4.20,1.95)
  -- (1.35,1.95) -- (1.35,2.19);
\draw[flow,black!65,densely dashed] (7.15,1.75) -- (7.15,2.19);

\end{tikzpicture}
\caption{\textbf{Method overview.} (i) GEE expands a small vocabulary of familiar glyphs into a large corpus of unfamiliar geometries paired with style-suppressed structural sketches. (ii) The Universal Geometry LoRA (UGL) is trained once on this corpus and then frozen, providing a general geometry-following prior.
(iii) Each Style LoRA is trained on top of the frozen UGL using only a few
style examples, with geometry reinforcement preserving its open-form capability.
(iv) At inference, UGL and the Style LoRA are applied jointly over the open-form inputs}.
\label{fig:overview}
\end{figure*}

\subsection{Structural Conditioning}

Spatially conditioned diffusion models provide a complementary line of related
work. ControlNet~\cite{zhang2023controlnet} and
T2I-Adapter~\cite{mou2024t2iadapter} condition pre-trained text-to-image models
on spatial signals such as edges, depth, segmentation, poses, and sketches,
while Sketch-Guided Diffusion~\cite{voynov2023sketch} focuses specifically on
generation controlled by free-form sketches. These methods demonstrate that
strong spatial adherence can be learned over broad image distributions.

Our problem, however, is not spatial conditioning alone. The structural sketch
specifies a form whose geometric realization must change according to a style
learned from a separate set of examples. The challenge is therefore to preserve
the target structure while preventing the learned style from carrying with it
the shape distribution on which that style was observed. We use structural
conditioning as a means toward this goal, learning a universal geometry prior
that remains fixed while different geometric styles are acquired. At inference,
we treat structure and style as independent conditions and adopt the
dual-conditioning guidance of InstructPix2Pix~\cite{instructpix2pix}, allowing
their strengths to be controlled separately.

\section{Method}
\label{sec:method}

Given a collection of shapes sharing a geometric style, our goal is to learn that style and apply it to a new target form specified by a structural sketch.
We use the term \emph{sketch-glyph} for a pair consisting of a structural sketch and its styled realization. 
The term originates naturally from fonts, where a sketch specifies the structure of a character and the glyph realizes it in a particular font. In this work, we extend this notion to pictograms,
symbols, and other structured forms that are not letters. 
Given a small collection of sketch-glyph pairs sharing a style, we aim to realize arbitrary new structural sketches in that style, as illustrated in Fig.~\ref{fig:open}.

Figure~\ref{fig:overview} provides an overview of the method. Our key design is to separate the two responsibilities involved in this task.
A \emph{Universal Geometry LoRA (UGL)} provides a general ability to faithfully follow the input structural sketch, while a \emph{Style LoRA} learns how the conditioned geometry should be realized in the target style.
The two adapters are learned hierarchically rather than independently. 
UGL is trained once (Fig.~\ref{fig:overview} \textit{ii}); each Style LoRA is subsequently trained on top of the frozen UGL (Fig.~\ref{fig:overview} \textit{iii}).

For this separation to support open-form transfer, UGL itself must generalize far beyond any particular shape family. We therefore train it on a large and purposefully diverse collection of synthetic geometries constructed by our \emph{Geometry Enrichment Engine (GEE)} (Fig.~\ref{fig:overview} \textit{i}). 

At inference (Fig.~\ref{fig:overview} \textit{iv}), UGL and the selected
Style LoRA are applied jointly. Their contributions can be controlled independently through
separate geometry and style guidance scales, allowing us to balance adherence
to the target structure against the strength of its stylistic realization.
The following sections describe UGL, GEE, and Style LoRA training in detail.

\subsection{Universal Geometry LoRA (UGL)}
\label{sec:ugl}

The purpose of UGL is to fine-tune a text-to-image diffusion model to follow open-form geometries (i.e. sketches).
Starting from a pre-trained text-to-image diffusion transformer~(DiT), we fit LoRA parameters on the DiT's weights using sketch-glyph data pairs constructed by GEE.
The rich, diverse synthetic data produced by the GEE allows us to select a high LoRA rank without overfitting, giving the UGL significant learning capacity.
After this one-time training, we freeze UGL and share it across all Style LoRAs, providing a common geometry prior beyond each style's specific training examples.

\subsection{Geometry Enrichment Engine (GEE)}
\label{sec:gee}

\input{figures/gee_overview}

Fonts are a natural starting point for data augmentation: they are available in a large variety of styles, and more importantly, their SVG native representation enables truly geometric augmentations compared to raster form.
However, using real font glyphs directly presents two limitations. First, available fonts cover a relatively narrow set of languages and therefore a limited vocabulary of structures, which may hinder generalization to unseen scripts and, even more, to out-of-domain forms such as symbols or pictograms.
Second, the geometry of a glyph is itself tied to the style of its source font, so the conditioning structure may carry unwanted stylistic cues. 
To address both issues, we introduce \emph{GEE}: a data-generation procedure that expands the range of training geometries while reducing their dependence on font style. 
To increase structural diversity, GEE takes advantage of the native SVG parameterization of fonts and constructs novel geometric forms by applying random sequences of geometric edits.
We refer to each such sequence as a \emph{geometric recipe}. 
A recipe may, for example, repeatedly remove or merge SVG parts, producing coherent but unfamiliar structures (Fig.~\ref{fig:GEM}(a)). 
Recipes also allow the GEE to address the second problem. A recipe applied to one font can be re-applied to matching letters from another font, producing paired data with \textbf{similar structure} but \textbf{different styles} (Fig.~\ref{fig:GEM}(b)).
During UGL training, we form mixed pairs by taking the conditioning sketch from one font and the target from another font to which the same recipe was applied (Fig.~\ref{fig:GEM}(c)). 
The sketch and target therefore remain aligned in structure while differing in style, encouraging UGL to learn the supplied geometry independently of its stylistic realization.

We next describe the editing primitives that drive the geometry enrichment process, first motivating their design and then outlining each one (full details in Appendix~\ref{sec:supp-gee}).

\paragraph*{Structural editing primitives.}
A glyph's structure is determined by how its strokes relate: which strokes meet and where they branch or terminate. Smooth transformations such as warping, shearing, or scaling can change a glyph's appearance, but they do not create new structural relations between its strokes. To generate genuinely new structures, we therefore need operations that modify these relations. We propose two elementary edits: component removal and merging. This builds on a broader principle: deleting and recombining existing parts can generate meaningful structural variation~\cite{funkhouser2004modeling}, and related cut-based augmentations have been shown to improve recognition performance~\cite{devries2017improved,Yun_2019_ICCV}. Repeating these edits produces novel structural relations beyond those found in existing alphabets.

To make both edits transferable across fonts, we edit the glyph's structure rather than its contour, which is more sensitive to stylistic variation. We therefore represent each glyph as a graph, with strokes as edges and intersections and endpoints as nodes. A recipe is then expressed as a sequence of graph edits, with component removal and merging each applied one to three times. The edit type and its structural parameters are sampled when the recipe is created, specifying which component to remove or where components should connect. Because the edits are defined structurally, the same recipe can be replayed across fonts to produce the same underlying structure in multiple styles. Structural correspondence across styles is further explained in Appendix~\ref{sec:supp-gee}.

\paragraph*{Graph extraction.}
\label{sec:gee_graph}
We represent each glyph as a graph that captures its dominant stroke structure (Fig.~\ref{fig:graph_extraction}). Starting from the glyph (a), we compute its medial axis together with the local inscribed-disk radius, which provides an estimate of stroke thickness (b). We prune short branches caused by serifs and other stylistic details, leaving the main stroke structure (c). Pixel chains are then collapsed into edges, with junctions and stroke endpoints represented as nod    es (d). Finally, we classify each edge by its structural role: an edge between two junctions is a \emph{bar}, while an edge from a junction to an endpoint is a \emph{leg} (e).
\usetikzlibrary{shadows}%
\begin{figure}[!h]
\centering
\providecommand{\gedir}{figures/assets/gee_graph_extraction}
\ifdefined\gecellw\else\newlength{\gecellw}\fi
\ifdefined\gecellh\else\newlength{\gecellh}\fi
\ifdefined\gegap\else\newlength{\gegap}\fi
\ifdefined\gevgap\else\newlength{\gevgap}\fi
\setlength{\gegap}{7pt}%
\setlength{\gecellw}{\dimexpr(\linewidth-4\gegap-1pt)/5\relax}%
\setlength{\gecellh}{1.145\gecellw}%
\setlength{\gevgap}{3pt}%
\providecommand{\geframe}{}\renewcommand{\geframe}{black!25}
\providecommand{\gecell}{}\renewcommand{\gecell}[4]{%
  \node[draw=\geframe,line width=.4pt,fill=white,
        drop shadow={shadow xshift=.8pt,shadow yshift=-.8pt,opacity=.25,fill=black},inner sep=0pt,
        minimum width=\gecellw,minimum height=\gecellh,anchor=north west]
    (#1) at ({#2*(\gecellw+\gegap)},{-#3*(\gecellh+\gevgap)})
    {\includegraphics[width=\dimexpr\gecellw-4pt\relax,height=\dimexpr\gecellh-4pt\relax,keepaspectratio]{\gedir/#4}};}
\providecommand{\gelab}{}\renewcommand{\gelab}[3]{%
  \node[anchor=south,inner sep=0pt,font=\comfortaa\fontsize{6pt}{7pt}\selectfont] at ({#1},{#2+2pt}) {(#3)};}
\begin{tikzpicture}[gearrow/.style={-{Straight Barb[length=2pt,width=2.6pt]},line width=.4pt,black!45,
                                    shorten <=1pt,shorten >=1pt}]
\foreach \r/\g in {0/row1_mu,1/row2_thorn} {
  \renewcommand{\geframe}{black}
  \gecell{g\r a}{0}{\r}{\g_1_glyph}
  \renewcommand{\geframe}{black!25}
  \gecell{g\r b}{1}{\r}{\g_2_mask}
  \gecell{g\r c}{2}{\r}{\g_4_prune}
  \gecell{g\r d}{3}{\r}{\g_5_edges}
  \gecell{g\r e}{4}{\r}{\g_6_cands}
  \foreach \a/\b in {a/b,b/c,c/d,d/e} \draw[gearrow] (g\r\a) -- (g\r\b);
}
\foreach \c/\l in {0/a,1/b,2/c,3/d,4/e} {\gelab{\c*(\gecellw+\gegap)+.5*\gecellw}{0pt}{\l}}
\end{tikzpicture}
\caption{\textbf{Graph extraction.} The steps of extracting a structural graph from a glyph, shown for two glyphs. Columns labelled as referenced in Sec.~\ref{sec:gee_internals}.}
\label{fig:graph_extraction}
\end{figure}
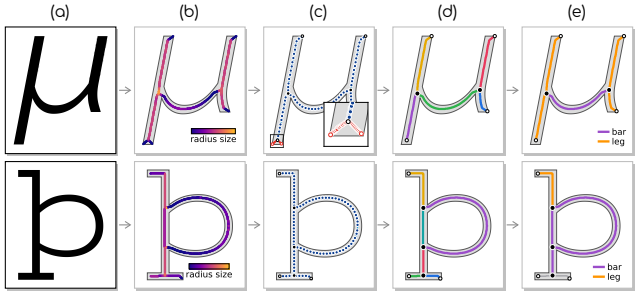

\usetikzlibrary{shadows}%
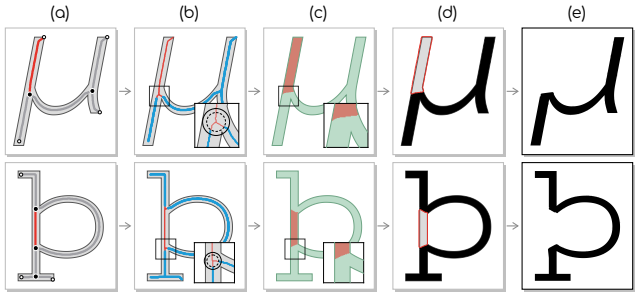
\begin{figure}[!b]
\centering
\providecommand{\cmdir}{figures/assets/gee_component_removal}
\ifdefined\cmcellw\else\newlength{\cmcellw}\fi
\ifdefined\cmcellh\else\newlength{\cmcellh}\fi
\ifdefined\cmgap\else\newlength{\cmgap}\fi
\ifdefined\cmvgap\else\newlength{\cmvgap}\fi
\setlength{\cmgap}{7pt}%
\setlength{\cmcellw}{\dimexpr(\linewidth-4\cmgap-1pt)/5\relax}%
\setlength{\cmcellh}{1.145\cmcellw}%
\setlength{\cmvgap}{3pt}%
\providecommand{\cmframe}{}\renewcommand{\cmframe}{black!25}
\providecommand{\cmcell}{}\renewcommand{\cmcell}[4]{%
  \node[draw=\cmframe,line width=.4pt,fill=white,
        drop shadow={shadow xshift=.8pt,shadow yshift=-.8pt,opacity=.25,fill=black},inner sep=0pt,
        minimum width=\cmcellw,minimum height=\cmcellh,anchor=north west]
    (#1) at ({#2*(\cmcellw+\cmgap)},{-#3*(\cmcellh+\cmvgap)})
    {\includegraphics[width=\dimexpr\cmcellw-4pt\relax,height=\dimexpr\cmcellh-4pt\relax,keepaspectratio]{\cmdir/#4}};}
\providecommand{\cmlab}{}\renewcommand{\cmlab}[3]{%
  \node[anchor=south,inner sep=0pt,font=\comfortaa\fontsize{6pt}{7pt}\selectfont] at ({#1},{#2+2pt}) {(#3)};}
\begin{tikzpicture}[cmarrow/.style={-{Straight Barb[length=2pt,width=2.6pt]},line width=.4pt,black!45,
                                    shorten <=1pt,shorten >=1pt}]
\foreach \r/\g in {0/row1_mu,1/row2_thorn} {
  \renewcommand{\cmframe}{black!25}
  \cmcell{c\r a}{0}{\r}{\g_7_select}
  \cmcell{c\r b}{1}{\r}{\g_8_ball}
  \cmcell{c\r c}{2}{\r}{\g_10_keep}
  \cmcell{c\r d}{3}{\r}{\g_11_sub}
  \renewcommand{\cmframe}{black}
  \cmcell{c\r e}{4}{\r}{\g_14_result}
  \foreach \a/\b in {a/b,b/c,c/d,d/e} \draw[cmarrow] (c\r\a) -- (c\r\b);
}
\foreach \c/\l in {0/a,1/b,2/c,3/d,4/e} {\cmlab{\c*(\cmcellw+\cmgap)+.5*\cmcellw}{0pt}{\l}}
\end{tikzpicture}
\caption{\textbf{Component removal.} The steps of removing one structural component, shown for two glyphs. Columns labelled as referenced in Sec.~\ref{sec:gee_internals}.}
\label{fig:component_removal}
\end{figure}

\paragraph*{Component removal.}
\label{sec:gee_internals}
Fig.~\ref{fig:component_removal} illustrates the process. Our strategy is to perform the edit in the medial-axis representation and then reconstruct the glyph back in the SVG domain. Outline points unrelated to the removed component are preserved exactly. Only the region surrounding the cut is reconstructed, where additional processing is needed to obtain a clean, style-consistent boundary. Using the graph extracted in the previous step, we locate the component to remove from its structure, which lets us find the same component across styles (a). As seen in Fig.~\ref{fig:component_removal}(b), simply removing the red branch can leave bulges where it connects to the remaining blue branches. We therefore trim these regions and reconnect the remaining blue branches. We then separate the pixels belonging to the removed region from those that should remain (c). This produces a pixel removal mask, which we convert to a smooth vector representation (d).%
 Additional reconstruction steps are described in the appendix. The final removal result is shown in (e).

\paragraph*{Merging.}
\label{sec:gee_merging}
Fig.~\ref{fig:merge_graph} illustrates the process. We begin with two glyphs from the same style: a \emph{guest} (a) and a \emph{host} (b), and aim to combine them into a single coherent glyph. As in component removal, we perform the edit in the medial-axis representation and reconstruct the result in the SVG domain. Using the graphs extracted in the previous step, we identify where the guest should connect to the host based on their structural attributes. The process starts by randomly generating a candidate merging recipe, combining a set of simple augmentations, such as mirroring, rotation, and scaling, with a pair of connection points selected from the guest and host graphs, shown as red points in (c,d). To assess the result produced by the recipe, we apply the augmentations and connect the glyphs at the selected points. We approve only recipes that produce a clean, coherent shape (e). A merging recipe is retained only if it succeeds across a minimum number of styles. Additional implementation details are provided in the appendix. The final merging result is shown in (f).\usetikzlibrary{shadows}%
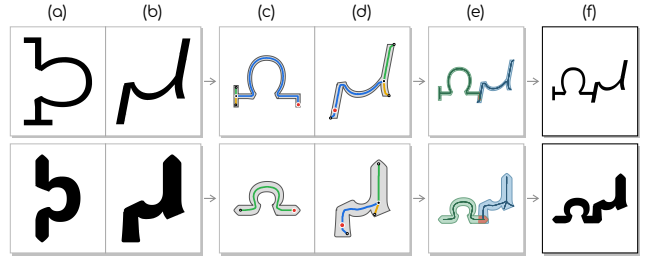
\begin{figure}[!h]
\centering
\providecommand{\mgdir}{figures/assets/gee_merging}
\ifdefined\mgcellw\else\newlength{\mgcellw}\fi
\ifdefined\mgcellh\else\newlength{\mgcellh}\fi
\ifdefined\mggap\else\newlength{\mggap}\fi
\ifdefined\mgvgap\else\newlength{\mgvgap}\fi
\setlength{\mggap}{7pt}%
\setlength{\mgcellw}{\dimexpr(\linewidth-3\mggap-1pt)/6\relax}%
\setlength{\mgcellh}{1.145\mgcellw}
\setlength{\mgvgap}{3pt}%
\providecommand{\mgframe}{}\renewcommand{\mgframe}{black!25}
\providecommand{\mgimg}{}\renewcommand{\mgimg}[1]{\includegraphics[width=\dimexpr\mgcellw-4pt\relax,
  height=\dimexpr\mgcellh-4pt\relax,keepaspectratio]{\mgdir/#1}}
\providecommand{\mgcell}{}\renewcommand{\mgcell}[4]{%
  \node[draw=\mgframe,line width=.4pt,fill=white,
        drop shadow={shadow xshift=.8pt,shadow yshift=-.8pt,opacity=.25,fill=black},inner sep=0pt,
        minimum width=\mgcellw,minimum height=\mgcellh,anchor=north west] (#1) at (#2,#3) {\mgimg{#4}};}
\providecommand{\mgpair}{}\renewcommand{\mgpair}[5]{%
  \node[draw=\mgframe,line width=.4pt,fill=white,
        drop shadow={shadow xshift=.8pt,shadow yshift=-.8pt,opacity=.25,fill=black},inner sep=0pt,
        minimum width=2\mgcellw,minimum height=\mgcellh,anchor=north west] (#1) at (#2,#3) {};
  \draw[black!35,line width=.25pt] (#1.north) -- (#1.south);
  \node[inner sep=0pt] at ([xshift=-.5\mgcellw]#1.center) {\mgimg{#4}};
  \node[inner sep=0pt] at ([xshift=.5\mgcellw]#1.center) {\mgimg{#5}};}
\providecommand{\mglab}{}\renewcommand{\mglab}[3]{%
  \node[anchor=south,inner sep=0pt,font=\comfortaa\fontsize{6pt}{7pt}\selectfont] at ({#1},{#2+2pt}) {(#3)};}
\begin{tikzpicture}[mgarrow/.style={-{Straight Barb[length=2pt,width=2.6pt]},line width=.4pt,black!45,
                                    shorten <=1pt,shorten >=1pt}]
\foreach \r/\g in {0/row1_stint,1/row2_adlam} {
  \pgfmathsetlengthmacro{\mgrowy}{-\r*(\mgcellh+\mgvgap)}
  \renewcommand{\mgframe}{black!25}
  \mgpair{m\r a}{0pt}{\mgrowy}{\g_2_guest}{\g_1_host}
  \mgpair{m\r b}{2\mgcellw+\mggap}{\mgrowy}{\g_5_graph_b}{\g_4_graph_a}
  \mgcell{m\r c}{4\mgcellw+2\mggap}{\mgrowy}{\g_6_gates}
  \renewcommand{\mgframe}{black}
  \mgcell{m\r d}{5\mgcellw+3\mggap}{\mgrowy}{\g_7_result}
  \draw[mgarrow] (m\r a) -- (m\r b);
  \draw[mgarrow] (m\r b) -- (m\r c);
  \draw[mgarrow] (m\r c) -- (m\r d);
}
\foreach \x/\l in {.5/a,1.5/b} {\mglab{\x*\mgcellw}{0pt}{\l}}
\foreach \x/\l in {2.5/c,3.5/d} {\mglab{\x*\mgcellw+\mggap}{0pt}{\l}}
\mglab{4.5*\mgcellw+2*\mggap}{0pt}{e}
\mglab{5.5*\mgcellw+3*\mggap}{0pt}{f}
\end{tikzpicture}
\caption{\textbf{Merging.} The steps of merging the same pair of glyphs in two styles. Columns labelled as referenced in Sec.~\ref{sec:gee_merging}.}
\label{fig:merge_graph}
\end{figure}

\subsection{Style LoRA}
\label{sec:style-lora}

We train a separate \emph{Style LoRA} on sketch-glyph pairs
for each target style, as shown in Fig.~\ref{fig:overview}(iii). The backbone
and UGL parameters remain fixed, while UGL continues to guide the geometry
throughout training. Only the Style LoRA is optimized, allowing it to focus on how that
geometry is realized in the target style. Here we use a low-rank adapter,
far smaller than UGL: its limited capacity helps keep the style update from
re-learning geometry or competing with the shared geometry prior.
Following previous work~\citep{gal2022image,ruiz2023dreambooth}, we
associate each style with a unique rare token. This token and a short
appearance description provide the textual conditioning during both Style
LoRA training and inference.
\paragraph*{Geometry reinforcement.}
The few style examples also share a limited family of geometries, so the
Style LoRA may still tie the learned style to those geometries despite
UGL's structural guidance. To preserve open-form transfer, we mix geometry-enriched
sketch-glyph pairs into its training batches, shown by the connection from
GEE to Style LoRA training in Fig.~\ref{fig:overview}. These pairs expose the
model to geometries outside the target style's examples throughout adaptation,
reinforcing its ability to follow new structures while learning the style.

\subsection{Inference}
\label{sec:inference}

After training, we apply UGL and the selected Style LoRA jointly to the base diffusion model to map structural sketches to styled outputs (Fig.~\ref{fig:overview} \textit{iv}). 
To control geometry adherence and style strength separately, we adopt dual classifier-free guidance following InstructPix2Pix~\cite{instructpix2pix}. 
During training, the structural sketch, the style caption, or both are dropped to their null conditions, so a single network learns all four conditioning combinations.
During sampling, we combine three forward passes using two CFG coefficients to obtain separate geometry and style guidance strengths, balancing adherence to the input sketch geometry vs.\ the target style.
Appendix~\ref{subsec:sup_dual_guidance} provides the guidance formulation and implementation details.

This joint application supports transfer to new forms without further training: a pictogram style can depict new human poses, while a sculptural style can turn free-form strokes into tubular shapes (Fig.~\ref{fig:teaser}). 
The same sketch can also be rendered in different learned styles (Fig.~\ref{fig:open}).
Section~\ref{sec:experiments} evaluates these capabilities through qualitative and quantitative comparisons and ablations of the method's main components.

\section{Experiments}
\label{sec:experiments}
We evaluate our method across fonts, pictograms, and object-centric styles, using sketches of increasing structural novelty:
\begin{itemize}
  \item \textbf{In-domain}, held-out test examples from the same source domain as the style training data
  \item \textbf{Near-domain}, semantically related but structurally unseen examples, such as non-Latin glyphs for fonts and everyday human poses for pictograms
  \item \textbf{Out-of-domain}, structures from unrelated shape families, such as non-glyph sketches for font styles or non-human sketches for pictogram styles.
\end{itemize}
We compare against prior methods and perform targeted ablations to assess both open-set transfer quality and the contributions of our main design choices.

\definecolor{gggreen}{RGB}{220,40,40}
\ifdefined\ggcell\else\newlength{\ggcell}\fi   %
\providecommand{\ggim}{}\renewcommand{\ggim}[1]{\includegraphics[width=\ggcell]{figures/assets/generalization_ablation/#1.jpg}}
\providecommand{\ggsks}{}\renewcommand{\ggsks}{0.7}   %
\providecommand{\ggsk}{}\renewcommand{\ggsk}[1]{\raisebox{\dimexpr(\ggcell-\ggsks\ggcell)/2}{%
  \includegraphics[width=\ggsks\ggcell]{figures/assets/generalization_ablation/#1.jpg}}}
\providecommand{\ggrefs}{}\renewcommand{\ggrefs}{0.8}
\providecommand{\ggref}{}\renewcommand{\ggref}[1]{\raisebox{\dimexpr(\ggcell-\ggrefs\ggcell)/2}{%
  \includegraphics[width=\ggrefs\ggcell]{figures/assets/generalization_ablation/#1.jpg}}}
\providecommand{\gggap}{}\renewcommand{\gggap}{7pt}
\providecommand{\ggrule}{}\renewcommand{\ggrule}{\hskip\gggap{\color{black!40}\vrule width 0.3pt}\hskip\gggap}
\providecommand{\gggaponly}{}\renewcommand{\gggaponly}{\hskip\gggap\hskip 0.3pt\hskip\gggap}
\providecommand{\ggrefimg}{}\renewcommand{\ggrefimg}[2]{\raisebox{\dimexpr(\ggcell-\ggrefs\ggcell)/2}{%
  \tikz[remember picture,baseline=(#2.south)]\node[inner sep=0pt,outer sep=0pt](#2){\includegraphics[width=\ggrefs\ggcell]{figures/assets/generalization_ablation/#1.jpg}};}}
\providecommand{\ggrefL}{}\renewcommand{\ggrefL}[2]{\ggrefimg{#1}{#2}}
\providecommand{\ggrefR}{}\renewcommand{\ggrefR}[2]{\ggrefimg{#1}{#2R}%
  \tikz[remember picture,overlay]\draw[dashed,line width=0.6pt]
    ([shift={(-3pt,-3pt)}]#2.south west) rectangle ([shift={(3pt,3pt)}]#2R.north east);}
\providecommand{\ggbox}{}\renewcommand{\ggbox}[4]{\draw[draw=red,line width=0.5pt] (#1,#2) rectangle (#3,#4);}
\providecommand{\ggout}{}\renewcommand{\ggout}[2]{%
  \begin{tikzpicture}[baseline=(i.south)]
    \node[inner sep=0pt,outer sep=0pt] (i) {\ggim{#1}};
    \begin{scope}[shift={(i.south west)},x=\ggcell,y=\ggcell]#2\end{scope}
  \end{tikzpicture}}

\begin{figure*}[!t]
\centering
\providecommand{\gglabel}{}\renewcommand{\gglabel}[1]{{\comfortaa\scriptsize #1}}
\setlength{\tabcolsep}{0.7pt}
\setlength{\ggcell}{\dimexpr(\linewidth-60pt)*10/126\relax}
\renewcommand{\arraystretch}{0}%
\begin{tabular}{@{}ccc@{\gggaponly}cccc@{\ggrule}cccc@{\ggrule}cccc@{}}
  \ggrefL{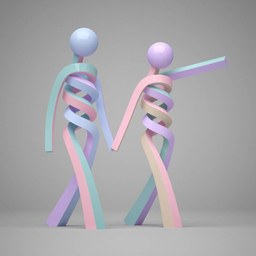}{ggrleesan} &
  \ggref{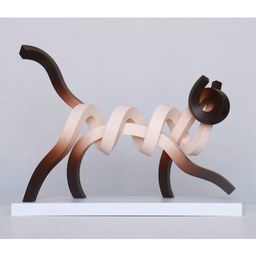} &
  \ggrefR{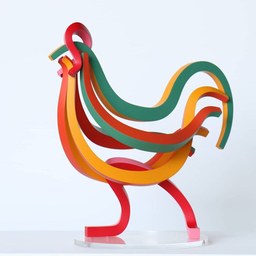}{ggrleesan} &
  \ggsk{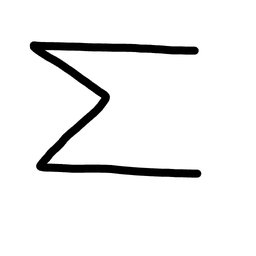} &
  \ggout{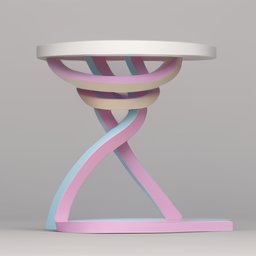}{\ggbox{0.099}{0.428}{0.888}{0.954}} &
  \ggsk{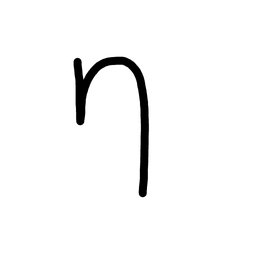} &
  \ggout{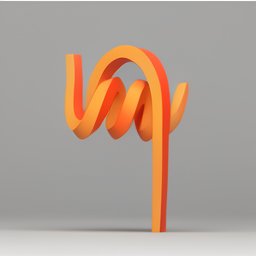}{} &
  \ggsk{sketch_sigma} &
  \ggout{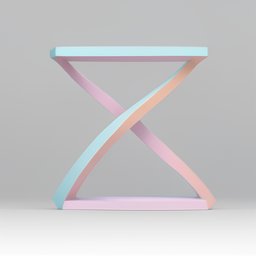}{\ggbox{0.454}{0.079}{0.908}{0.928}} &
  \ggsk{sketch_eta} &
  \ggout{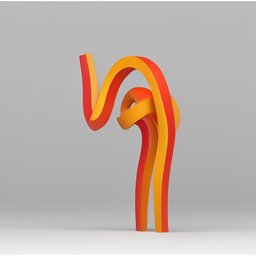}{} &
  \ggsk{sketch_sigma} &
  \ggim{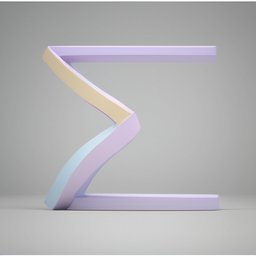} &
  \ggsk{sketch_eta} &
  \ggim{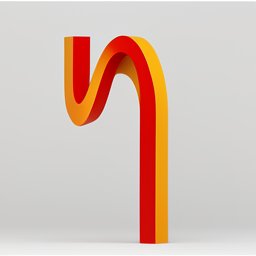} \\[1.5pt]
  \ggrefL{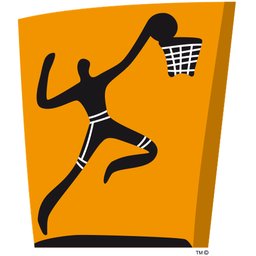}{ggrathens} &
  \ggref{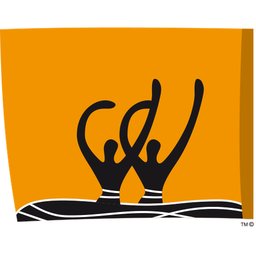} &
  \ggrefR{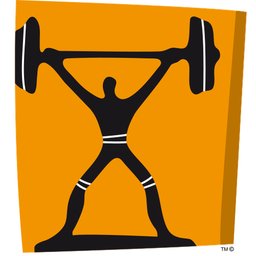}{ggrathens} &
  \ggsk{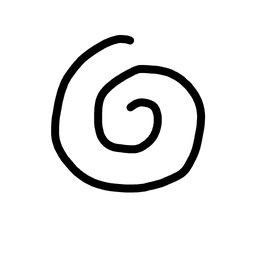} &
  \ggout{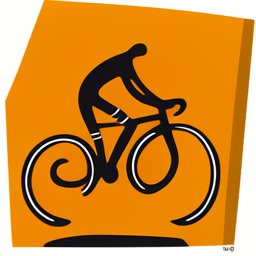}{} &
  \ggsk{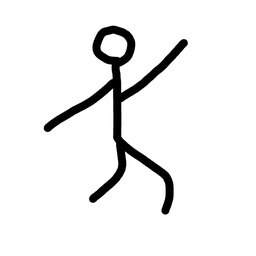} &
  \ggout{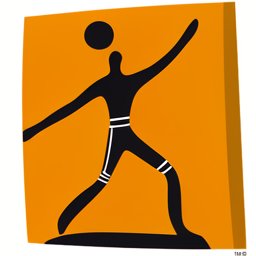}{\ggbox{0.267}{0.517}{0.587}{0.971}} &
  \ggsk{sketch_spiral} &
  \ggout{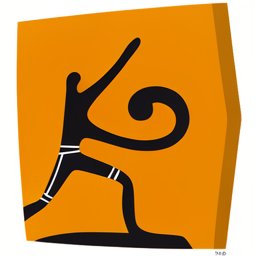}{\ggbox{0.114}{0.08}{0.426}{0.864}} &
  \ggsk{sketch_figure} &
  \ggout{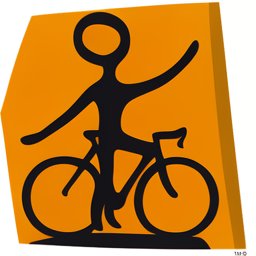}{} &
  \ggsk{sketch_spiral} &
  \ggim{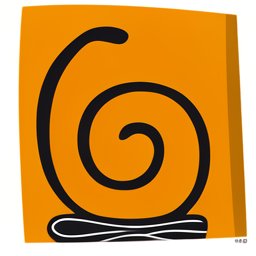} &
  \ggsk{sketch_figure} &
  \ggim{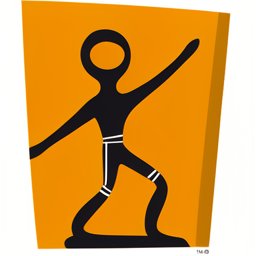} \\[1.5pt]
  \ggrefL{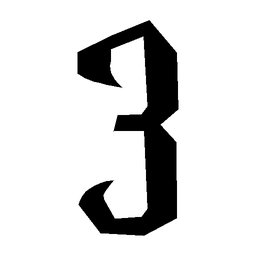}{ggrharryp} &
  \ggref{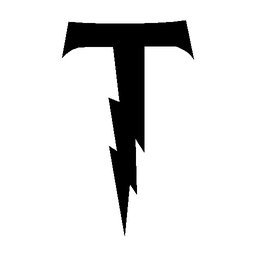} &
  \ggrefR{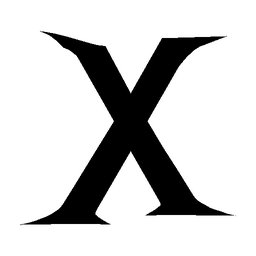}{ggrharryp} &
  \ggsk{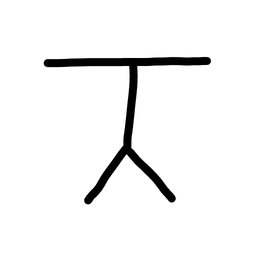} &
  \ggout{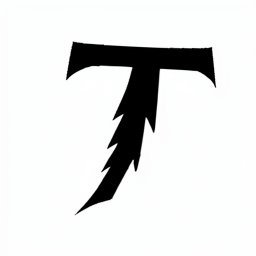}{} &
  \ggsk{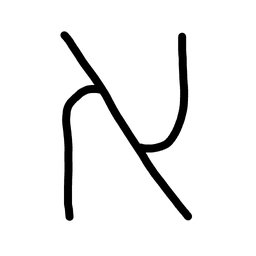} &
  \ggout{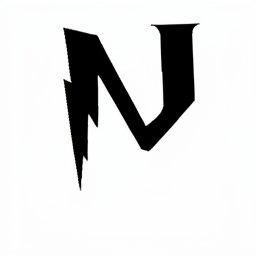}{} &
  \ggsk{sketch_tee} &
  \ggout{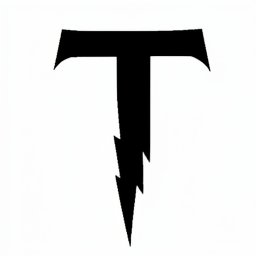}{} &
  \ggsk{sketch_aleph} &
  \ggout{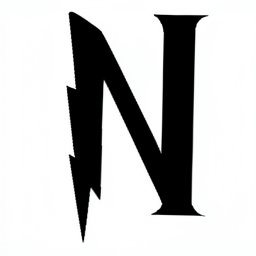}{} &
  \ggsk{sketch_tee} &
  \ggim{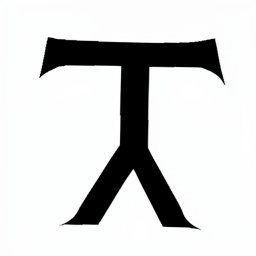} &
  \ggsk{sketch_aleph} &
  \ggim{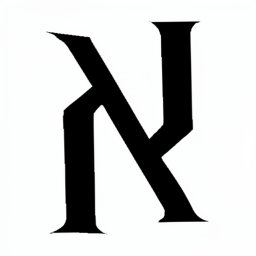} \\[4pt]
\multicolumn{3}{c}{\gglabel{Style Refs}} &
\multicolumn{4}{c}{\gglabel{w/o UGL}} &
\multicolumn{4}{c}{\gglabel{w/o GEE}} &
\multicolumn{4}{c}{\gglabel{Full Method (Ours)}} \\
\end{tabular}
\caption{\textbf{Path to geometric generalization.} Each pair shows an unseen sketch and the output in different styles.
\emph{w/o UGL} trains the Style LoRA without UGL, and \emph{w/o GEE} trains the UGL without GEE. 
In both, the failure patterns of the geometry are marked.
Ours follows the same sketches while preserving each style.}
\label{fig:geomgen-failure}
\end{figure*}

\subsection{Training and Evaluation Data}
\label{subsec:data}

The training data for the universal geometry LoRA (UGL) is produced by the geometry enrichment engine (GEE) from $80$ base glyphs (Latin letters, Latin Extended letters, and digits) shared across $16$ fonts. Applying $6{,}000$ random recipes yields $42{,}705$ sketch--glyph pairs, on which UGL is trained once (Appendix~\ref{subsec:training_details}). Each style is represented by a small exemplar set with manually drawn structural sketches. We hold out six examples per style for evaluation and train the Style LoRA on the remainder, yielding roughly 20--70 sketch--exemplar pairs per style. During Style LoRA training, $10\%$ of each batch is replayed from the UGL training pairs using their original style-free captions, helping preserve the learned geometry prior while adapting to style. For evaluation, we use a benchmark of $50$ curated sketches per style category (fonts and pictograms). To analyze how well our method and the baselines generalize, the sketches span the in-, near-, and out-of-domain tiers defined above. We evaluate across twelve styles, and for each style all methods receive the same sketches and style references.

\subsection{Open-form Geometric Generalization}
\label{sec:geometric-generalization}

We tailored our main technical contributions, the universal geometry LoRA (UGL) and the geometry enrichment engine (GEE), to open-form generalization and the challenges of fine-tuning large models. 

In this section, we demonstrate the necessity of both components; later in Table~\ref{tab:quantitative} we also present a quantitative evaluation that supports each claim.

We begin with the most direct alternative: avoid the UGL by fine-tuning a single low-rank adapter on the few sketch-glyph pairs available for a given style.  While this formulation preserves both the target geometry and the learned style on held-out examples from the \textbf{same style family}, it does not generalize to \textbf{out-of-domain} structures. Instead, the model takes a shortcut by producing results resembling the closest training example, best seen in the bottom rows of Fig.~\ref{fig:geomgen-failure} (\emph{w/o UGL}), where the model outputs styled examples similar to English alphabet letters rather than the desired geometry.
Row \emph{w/o UGL} in Table~\ref{tab:quantitative} quantifies this behavior.

\begin{figure}[!b]
    \centering
    \setlength{\tabcolsep}{1pt}
    \ifdefined\imw\else\newlength{\imw}\fi
    \setlength{\imw}{0.128\columnwidth}
    \providecommand{\im}[1]{\includegraphics[width=\imw]{#1}}
    \providecommand{\lab}{}\renewcommand{\lab}[1]{\makebox[0pt][c]{#1}}
    \providecommand{\refbox}{}\renewcommand{\refbox}[1]{%
        \tikz\node[draw=black!45, dashed, line width=0.4pt, inner xsep=3pt, inner ysep=1pt]{%
        \includegraphics[width=\dimexpr\imw-1.8pt]{figures/assets/qualitative_styles/#1_1.jpg}\hspace{1pt}%
        \includegraphics[width=\dimexpr\imw-1.8pt]{figures/assets/qualitative_styles/#1_2.jpg}};}
    \tikzset{figsep/.style={black!40, line width=0.3pt}}
    \providecommand{\ourslab}{}\renewcommand{\ourslab}{\lab{\bfseries\pdfliteral direct{2 Tr 0.25 w}Ours\pdfliteral direct{0 Tr 0 w}}}
    \providecommand{\mk}{}\renewcommand{\mk}[2]{\tikz[remember picture,overlay]\coordinate(#1) at (0,#2);}
    {\fontsize{6pt}{7.5pt}\selectfont\mdseries\comfortaa
    \noindent\makebox[\columnwidth][c]{%
        \begin{tabular*}{\columnwidth}{@{}c@{\extracolsep{\fill}}c@{\extracolsep{\fill}}c@{\extracolsep{\fill}}c@{\extracolsep{\fill}}c@{\extracolsep{\fill}}c@{}}%
            \mk{ftop}{\imw}\im{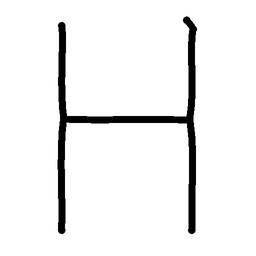} &%
            \refbox{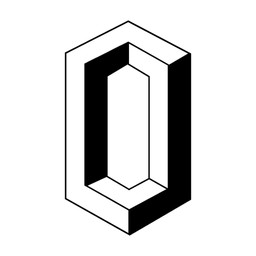}\mk{fg0R}{0pt} &%
            \mk{fg0L}{0pt}\im{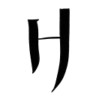} &%
            \im{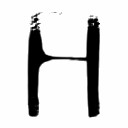}\mk{fg1R}{0pt} &%
            \mk{fg1L}{0pt}\im{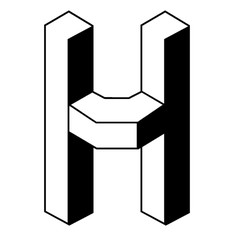} &%
            \im{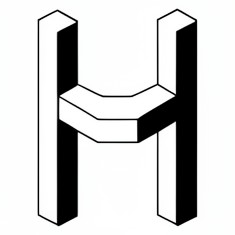} \\[5pt]%
            \mk{fbot}{-1pt}\im{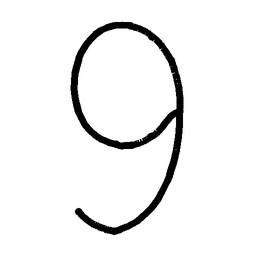} &%
            \refbox{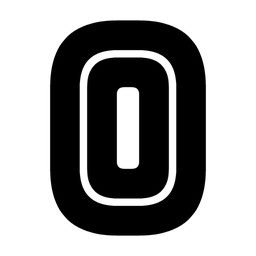} &%
            \im{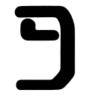} &%
            \im{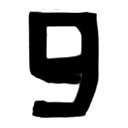} &%
            \im{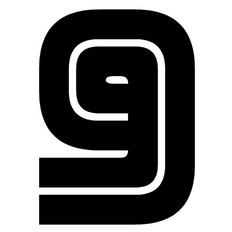} &%
            \im{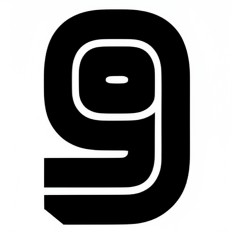} \\[-2pt]%
            \lab{Sketch} & \lab{Style Refs} & \lab{FontDiff.} & \lab{MX-Font} & \lab{GT} & \ourslab%
        \end{tabular*}}\par
    \vspace{4pt}\noindent\tikz\draw[figsep] (0,0) -- (\columnwidth,0);\par\vspace{4pt}
    \noindent\makebox[\columnwidth][c]{%
        \begin{tabular*}{\columnwidth}{@{}c@{\extracolsep{\fill}}c@{\extracolsep{\fill}}c@{\extracolsep{\fill}}c@{\extracolsep{\fill}}c@{\extracolsep{\fill}}c@{}}%
            \mk{ctop}{\imw}\im{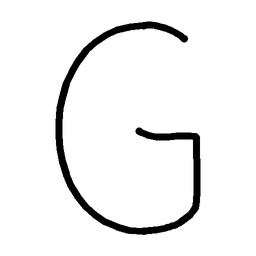} &%
            \refbox{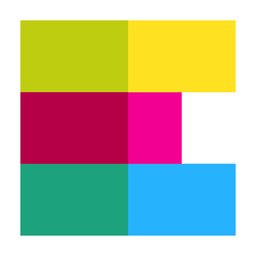}\mk{cg0R}{0pt} &%
            \mk{cg0L}{0pt}\im{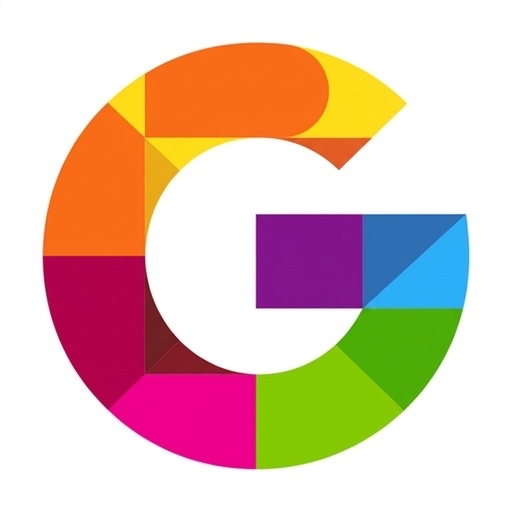} &%
            \im{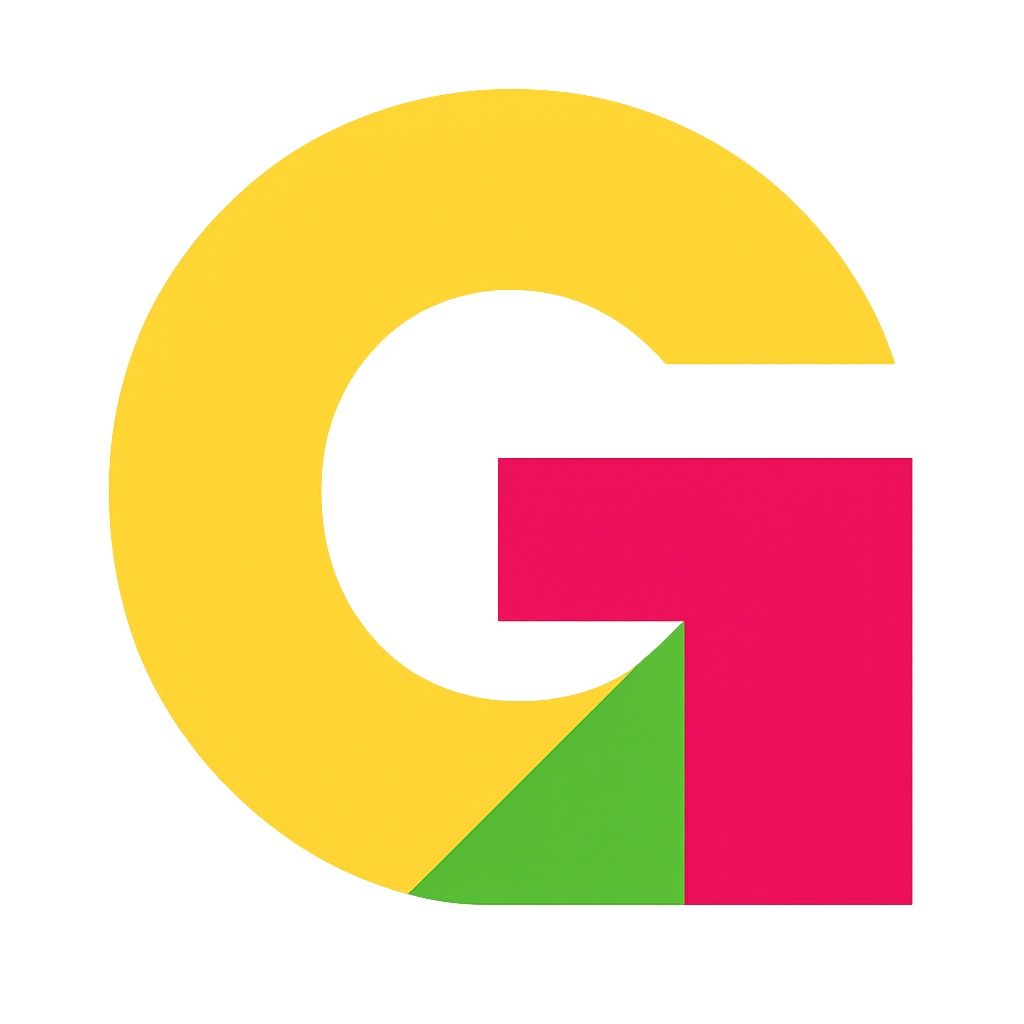}\mk{cg1R}{0pt} &%
            \mk{cg1L}{0pt}\im{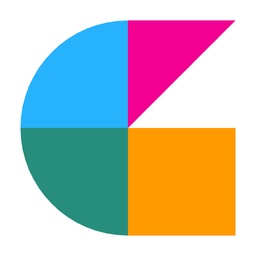} &%
            \im{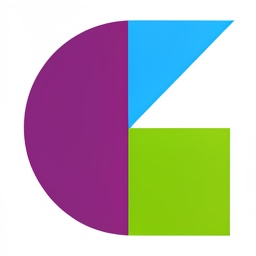} \\[5pt]%
            \im{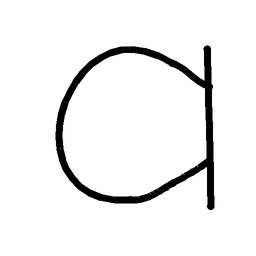} &%
            \refbox{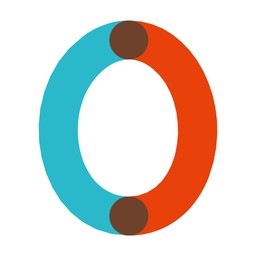} &%
            \im{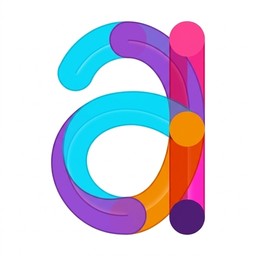} &%
            \im{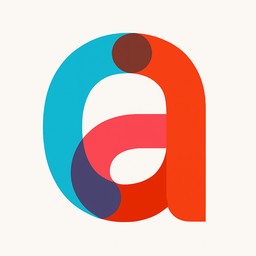} &%
            \im{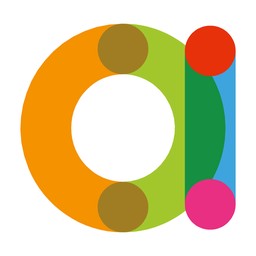} &%
            \im{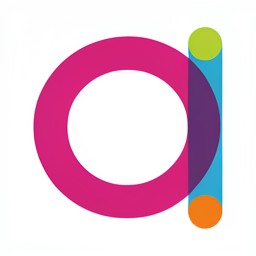} \\[5pt]%
            \mk{cbot}{-1pt}\im{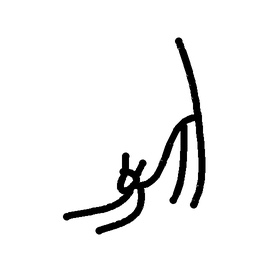} &%
            \refbox{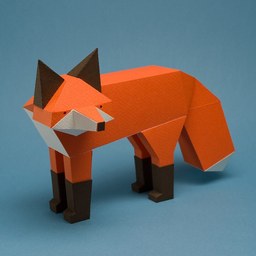} &%
            \im{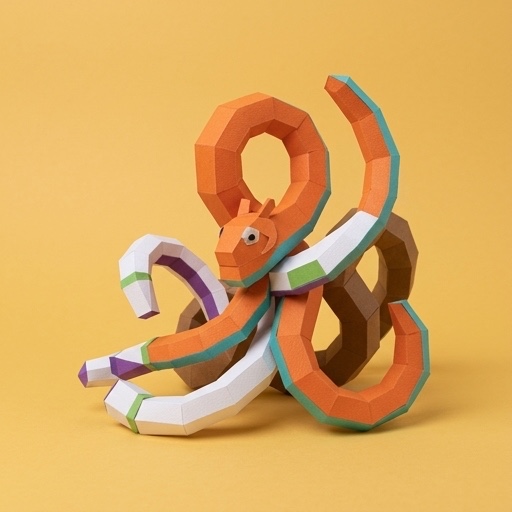} &%
            \im{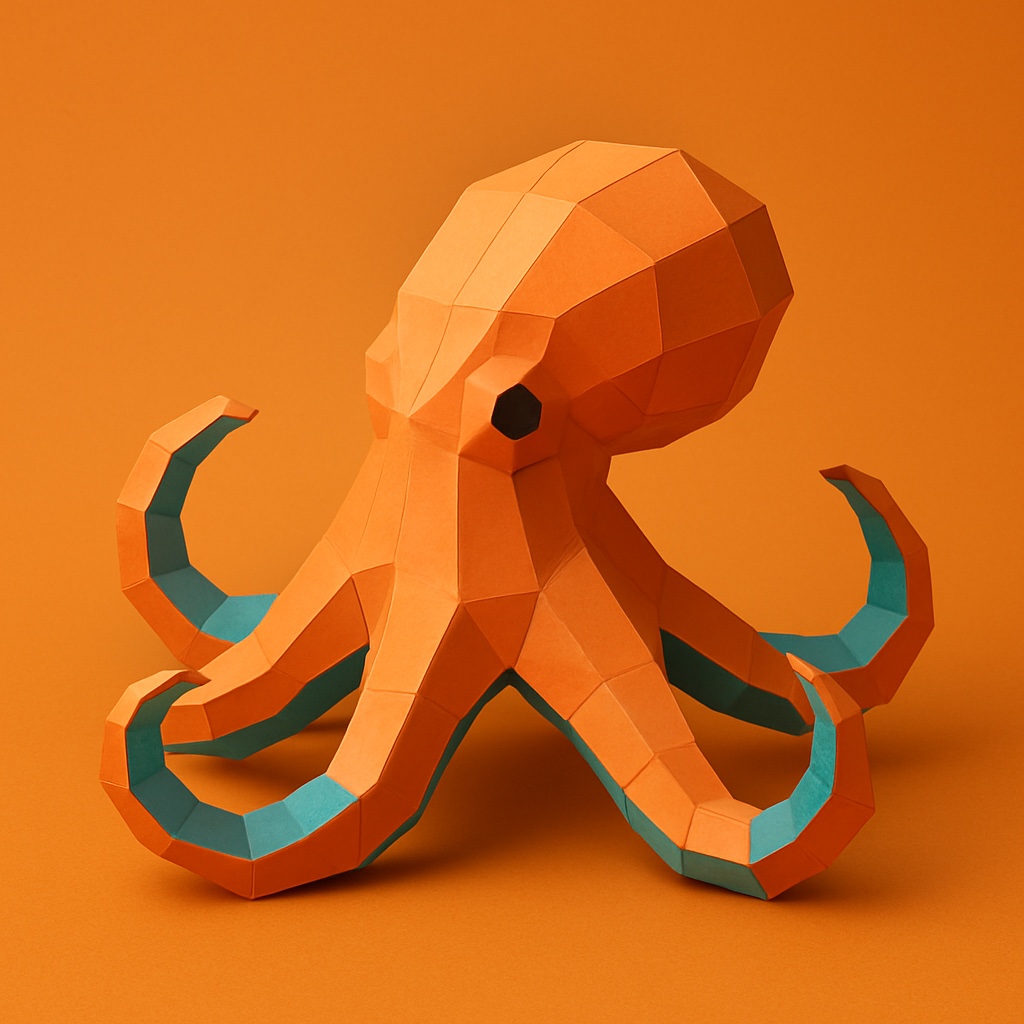} &%
            \im{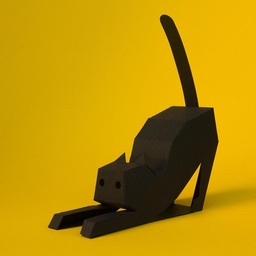} &%
            \im{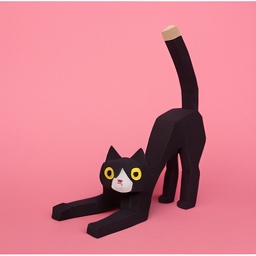} \\[-2pt]%
            \lab{Sketch} & \lab{Style Refs} & \lab{Gemini} & \lab{GPT} & \lab{GT} & \ourslab%
        \end{tabular*}}\par
    \tikz[remember picture,overlay]{%
      \coordinate (fv1) at ($(fg1R)!0.5!(fg1L)$); \draw[figsep] (fv1 |- ftop) -- (fv1 |- fbot);
      \coordinate (cv1) at ($(cg1R)!0.5!(cg1L)$); \draw[figsep] (cv1 |- ctop) -- (cv1 |- cbot);}%
    }
    \caption{\textbf{In-domain comparisons.} Each row shows a sketch held out from that style's training examples, two reference examples of the style, the outputs of the baselines, the ground truth, and our output. The font styles (top) are compared with the dedicated font-generation models, the colorful styles (bottom) with the general-purpose image models.}
    \label{fig:qual_fonts_in}
\end{figure}

A natural hypothesis for this phenomenon is that each geometry is observed only through a single realization in the target style, providing too little variation to disentangle the geometry and style. 

An intuitive solution is to train a geometry adapter on ordinary glyphs drawn from a broad corpus of fonts, using the same training settings as UGL but without employing GEE, and then train the same Style LoRAs on top of it. 
Unfortunately, the failure persisted, now in an even more revealing form. 
The UGL is even more biased towards common geometries from the train set, e.g. alphabet letters, as again seen in the bottom row of the \emph{w/o GEE} section of Fig~\ref{fig:geomgen-failure}.
These results are again supported by data in Table~\ref{tab:quantitative}.

What is instead needed is a training distribution with fewer geometric correlations, preventing the model from capturing a specific shape identity.
This is solved by our second technical contribution (GEE Sec.~\ref{sec:gee}). 
The GEE's immense augmentation strategy forces the geometry model to rely on the supplied structure rather than recognizing a known geometry. 
With this enriched corpus, completing our full method, the out-of-domain score rises to $0.94$ (\emph{Ours} in Table~\ref{tab:quantitative}), with a geometry factor of $0.97$ and style within 3\% of the ceiling, and the characteristic substitutions largely disappear.

\subsection{Qualitative Evaluation}
\label{subsec:exp_qual_eval}
We compare against FontDiffuser~\cite{fontdiffuser}, MX-Font~\cite{park2021mxfont}, Gemini 3.1 (Nano-Banana 2)~\cite{nanobanana2}, and gpt-image-1-mini~\cite{gptimage1mini}. All methods receive the same target sketch and style references, except FontDiffuser, which uses a single reference. We include the general-purpose models for completeness, as they require no per-style training. Additional comparisons, including the adapter-based methods B-LoRA and ZipLoRA as well as AiS, are provided in Appendix~\ref{sec:supp_further}.

\providecommand{\showdir}{figures/assets/qualitative_out_of_domain}
\begin{figure*}[h]
    \centering
    \ifdefined\shh\else\newlength{\shh}\fi
    \setlength{\shh}{0.108\textwidth}      %
    \providecommand{\shcell}{}\renewcommand{\shcell}[1]{%
        \begin{tabular}[b]{@{}c@{}}
            {\setlength{\fboxsep}{0pt}\setlength{\fboxrule}{0.4pt}%
             \fbox{\includegraphics[height=0.5\shh]{\showdir/#1_sketch.jpg}}} \\[1pt]
            \includegraphics[height=\shh]{\showdir/#1_res.jpg}
        \end{tabular}}
    \providecommand{\shref}{}\renewcommand{\shref}{%
        \begin{tabular}[b]{@{}c@{\hspace{3pt}}c@{}}
            \rotatebox{90}{\small\comfortaa Reference} &
            \tikz\node[draw, dashed, line width=0.6pt, inner sep=2pt]{%
                \begin{tabular}{@{}c@{}}
                    \includegraphics[height=0.74\shh]{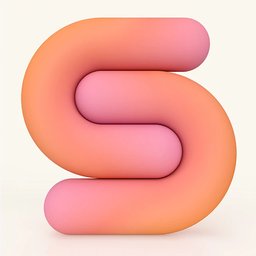} \\[1pt]
                    \includegraphics[height=0.74\shh]{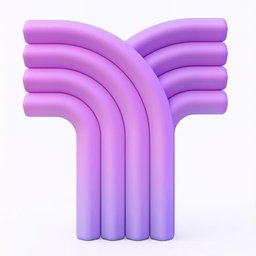}
                \end{tabular}};
        \end{tabular}}
    \providecommand{\shvsep}{}\renewcommand{\shvsep}{%
        {\color{black!30}\rule[-2pt]{0.8pt}{1.58\shh}}}
    \noindent\makebox[\textwidth]{\shref\hspace{8pt}\shcell{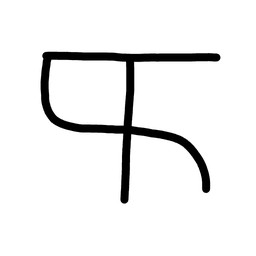}\hspace{2pt}\shcell{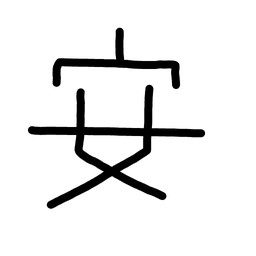}\hspace{2pt}\shcell{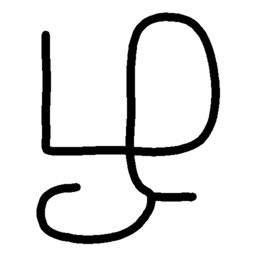}\hspace{2pt}\shcell{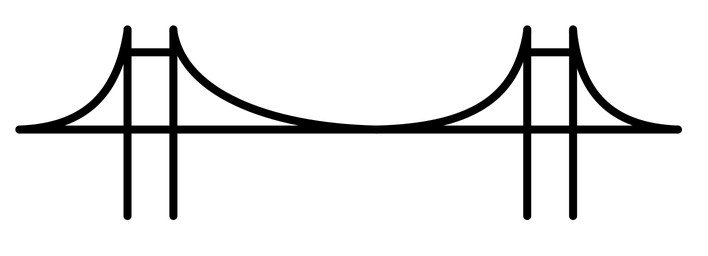}\hspace{2pt}\shcell{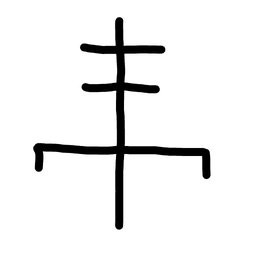}\hspace{2pt}\shcell{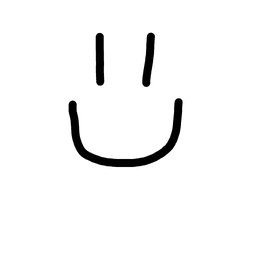}}\par
    \par\nointerlineskip\vspace{2pt}%
    {\color{black!30}\rule{0.97\linewidth}{0.8pt}}\par\nointerlineskip\vspace{4pt}%
    \noindent\makebox[\textwidth]{\shcell{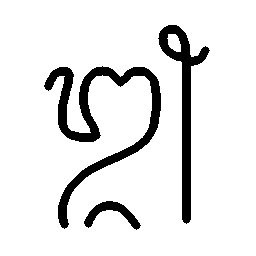}\hspace{2pt}\shcell{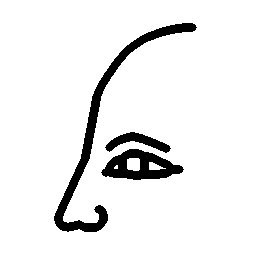}\hspace{5pt}\shvsep\hspace{5pt}\shcell{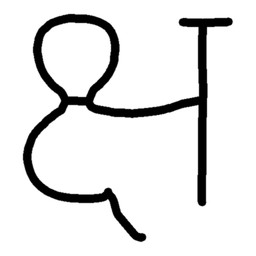}\hspace{2pt}\shcell{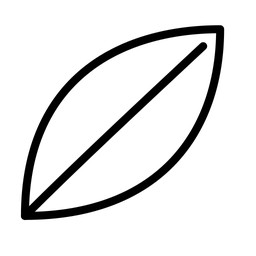}\hspace{2pt}\shcell{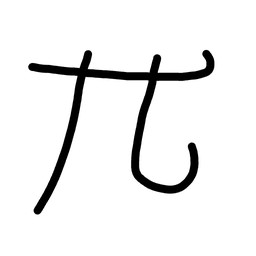}\hspace{2pt}\shcell{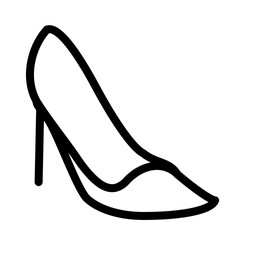}\hspace{5pt}\shvsep\hspace{5pt}\shcell{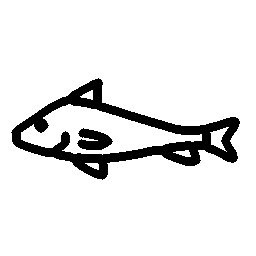}\hspace{2pt}\shcell{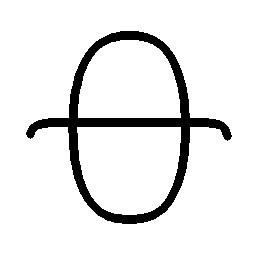}}
    \caption{\textbf{Out-of-domain results.} Each result is shown directly below its framed structural sketch. Each sketch lies outside the shape family used to learn the corresponding style. \textbf{Top row:} a single sculptural style applied to glyphs from unseen scripts, a bridge, and free-form shapes. Reference examples are included to illustrate the diversity within the learned style. \textbf{Bottom row:} three additional styles: two pictogram styles (left \& right) and a second sculptural style (middle)}
    \label{fig:showcase}

\end{figure*}

In the \emph{in-domain} tier, we evaluate held-out sketches (Fig.~\ref{fig:qual_fonts_in}). Our method closely matches the ground truth while preserving the requested geometry and fine stylistic details. Gemini and GPT often capture the overall style but drift toward familiar poses or glyphs and lose fine font details. FontDiffuser often deviates from the input structure, while MX-Font follows it more closely but transfers the target style less faithfully.

\begin{figure}[!h]
    \centering
    \setlength{\tabcolsep}{1pt}
    \ifdefined\imw\else\newlength{\imw}\fi
    \setlength{\imw}{0.119\columnwidth}
    \providecommand{\im}[1]{\includegraphics[width=\imw]{#1}}
    \providecommand{\lab}{}\renewcommand{\lab}[1]{\makebox[0pt][c]{#1}}
    \providecommand{\refbox}{}\renewcommand{\refbox}[1]{%
        \tikz\node[draw=black!45, dashed, line width=0.4pt, inner xsep=3pt, inner ysep=1pt]{%
        \includegraphics[width=\dimexpr\imw-1.8pt]{figures/assets/qualitative_styles/#1_1.jpg}\hspace{1pt}%
        \includegraphics[width=\dimexpr\imw-1.8pt]{figures/assets/qualitative_styles/#1_2.jpg}};}
    \tikzset{figsep/.style={black!40, line width=0.3pt}}
    \providecommand{\ourslab}{}\renewcommand{\ourslab}{\lab{\bfseries\pdfliteral direct{2 Tr 0.25 w}Ours\pdfliteral direct{0 Tr 0 w}}}
    \providecommand{\mk}{}\renewcommand{\mk}[2]{\tikz[remember picture,overlay]\coordinate(#1) at (0,#2);}
    \ifdefined\zw\else\newlength{\zw}\fi
    \setlength{\zw}{1.4\imw}
    \providecommand{\zoombox}{}\renewcommand{\zoombox}[4]{%
        \raisebox{\dimexpr(\imw-\zw)/2}{\tikz\node[draw=red, line width=0.5pt, inner sep=0.3pt]
        {\includegraphics[viewport=\fpeval{#2*256} \fpeval{#3*256} \fpeval{(#2+#4)*256} \fpeval{(#3+#4)*256}, clip, width=\dimexpr\zw-1.1pt]{#1}};}}
    \providecommand{\ourscrop}{}\renewcommand{\ourscrop}[4]{%
        \tikz[baseline=(o.south)]{%
        \node[inner sep=0pt] (o) {\includegraphics[width=\imw]{#1}};
        \draw[red, line width=0.4pt] ([shift={(#2\imw,#3\imw)}]o.south west) rectangle ++(#4\imw,#4\imw);}}
    {\fontsize{6pt}{7.5pt}\selectfont\mdseries\comfortaa
    \noindent\makebox[\columnwidth][c]{%
        \begin{tabular*}{\columnwidth}{@{}c@{\extracolsep{\fill}}c@{\extracolsep{\fill}}c@{\extracolsep{\fill}}c@{\extracolsep{\fill}}c@{\extracolsep{\fill}}c@{}}%
            \mk{ftop}{\imw}\im{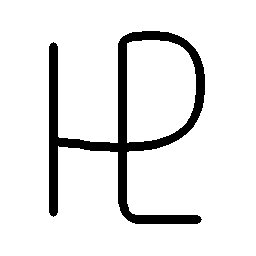} &%
            \refbox{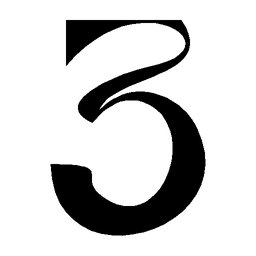}\mk{fg0R}{0pt} &%
            \mk{fg0L}{0pt}\im{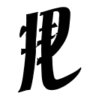} &%
            \im{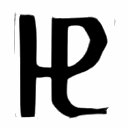}\mk{fg1R}{0pt} &%
            \mk{fg1L}{0pt}\ourscrop{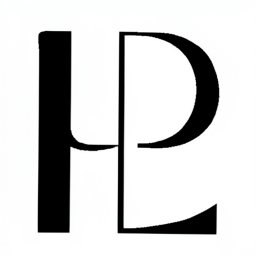}{0.407}{0.055}{0.481} &%
            \zoombox{figures/assets/qualitative_near_domain/font_1/ours.jpg}{0.407}{0.055}{0.481} \\[5pt]%
            \mk{fbot}{-1pt}\im{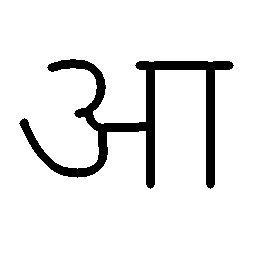} &%
            \refbox{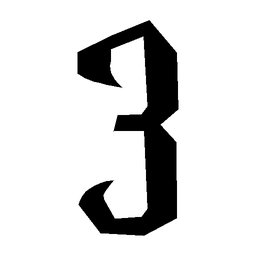} &%
            \im{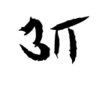} &%
            \im{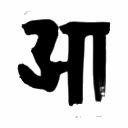} &%
            \ourscrop{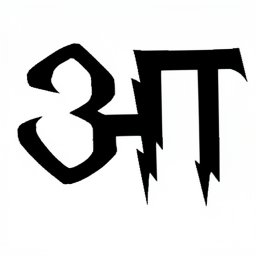}{0.498}{0.177}{0.406} &%
            \zoombox{figures/assets/qualitative_near_domain/font_4/ours.jpg}{0.498}{0.177}{0.406} \\[-2pt]%
            \lab{Sketch} & \lab{Style Refs} & \lab{FontDiff.} & \lab{MX-Font} & \ourslab & %
        \end{tabular*}}\par
    \vspace{4pt}\noindent\tikz\draw[figsep] (0,0) -- (\columnwidth,0);\par\vspace{4pt}
    \noindent\makebox[\columnwidth][c]{%
        \begin{tabular*}{\columnwidth}{@{}c@{\extracolsep{\fill}}c@{\extracolsep{\fill}}c@{\extracolsep{\fill}}c@{\extracolsep{\fill}}c@{\extracolsep{\fill}}c@{}}%
            \mk{ctop}{\imw}\im{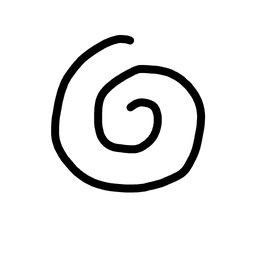} &%
            \refbox{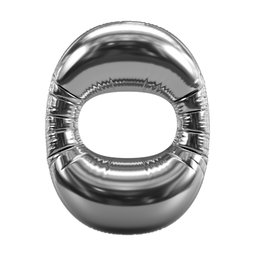}\mk{cg0R}{0pt} &%
            \mk{cg0L}{0pt}\im{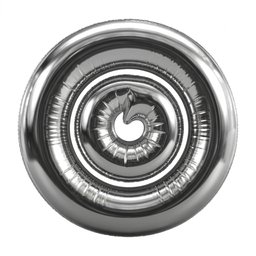} &%
            \im{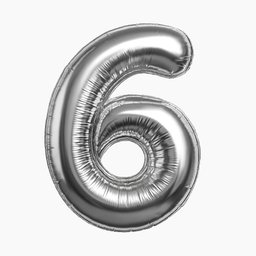}\mk{cg1R}{0pt} &%
            \mk{cg1L}{0pt}\ourscrop{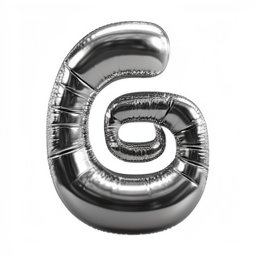}{0.558}{0.373}{0.283} &%
            \zoombox{figures/assets/qualitative_near_domain/puff/ours.jpg}{0.558}{0.373}{0.283} \\[5pt]%
            \mk{cbot}{-1pt}\im{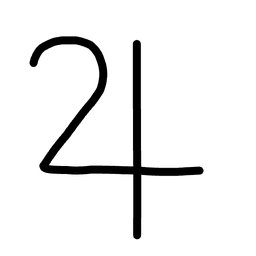} &%
            \refbox{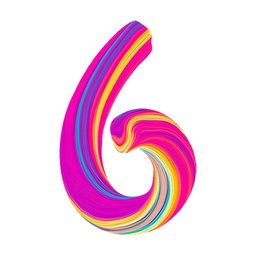} &%
            \im{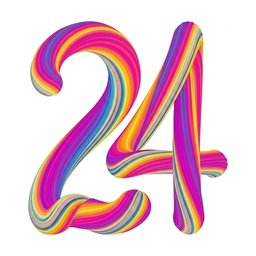} &%
            \im{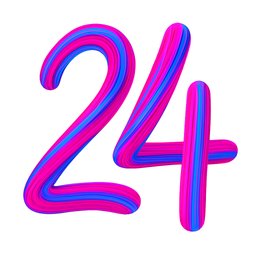} &%
            \ourscrop{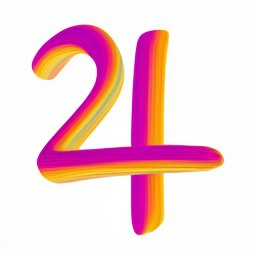}{0.107}{0.529}{0.39} &%
            \zoombox{figures/assets/qualitative_near_domain/candy/ours.jpg}{0.107}{0.529}{0.39} \\[-2pt]%
            \lab{Sketch} & \lab{Style Refs} & \lab{Gemini} & \lab{GPT} & \ourslab & %
        \end{tabular*}}\par
    \tikz[remember picture,overlay]{%
      \coordinate (fv1) at ($(fg1R)!0.5!(fg1L)$); \draw[figsep] (fv1 |- ftop) -- (fv1 |- fbot);
      \coordinate (cv1) at ($(cg1R)!0.5!(cg1L)$); \draw[figsep] (cv1 |- ctop) -- (cv1 |- cbot);}%
    }
    \caption{\textbf{Near-domain results.} Each row shows a non-Latin glyph sketch, two reference examples of the style, the outputs of the baselines, and our output next to an enlarged view of the region marked in red. The font styles (top) are compared with the dedicated font-generation models, the colorful styles (bottom) with the general-purpose image models.}
    \label{fig:qual_fonts_near}
\end{figure}

In the \emph{near-domain} tier (Fig.~\ref{fig:qual_fonts_near}), we transfer styles to related but unseen forms: non-Latin glyphs. Baselines often distort these unfamiliar inputs toward known forms or lose structural or stylistic details, while our method follows the supplied geometry and preserves the target style. The zoomed examples highlight reference motifs retained in the outputs, demonstrating transfer beyond the forms seen during style adaptation.

The \emph{out-of-domain} tier (Fig.~\ref{fig:showcase}) provides a stronger test by applying pictogram and object-centric styles to structures from unrelated shape families. Across these sketches, our method retains the characteristic shapes and appearance of each style while following the new structure, demonstrating open-form transfer beyond the source domain.

\subsection{Quantitative Results}
\begin{figure*}[!t]
\begin{minipage}[t]{0.64\textwidth}
\vspace{0pt}
\centering
\footnotesize
\setlength{\tabcolsep}{4pt}
\resizebox{\linewidth}{!}{%
\begin{tabular}[t]{l cc cc cc cc}
\toprule
 & \multicolumn{4}{c}{\textbf{VLM-as-a-judge}$\uparrow$} & \multicolumn{2}{c}{\textbf{LPIPS}$\downarrow$} & \multicolumn{2}{c}{\textbf{SSIM}$\uparrow$} \\
\cmidrule(lr){2-5} \cmidrule(lr){6-7} \cmidrule(lr){8-9}
 & \multicolumn{2}{c}{Per-style} & \multicolumn{2}{c}{Distance from domain} & & & & \\
\cmidrule(lr){2-3} \cmidrule(lr){4-5}
\textbf{Method} & Fonts & Non-Fonts & Near-domain & Out-of-domain & Fonts & Non-Fonts & Fonts & Non-Fonts \\
\midrule
FontDiffuser             & 0.459 & NA    & 0.469 & 0.433 & 0.425 & NA    & 0.473 & NA    \\
MX-Font                  & 0.597 & NA    & 0.627 & 0.522 & 0.315 & NA    & 0.508 & NA    \\
GPT-Image-1-mini         & 0.773 & 0.713 & 0.766 & 0.796 & 0.299 & 0.304 & 0.473 & 0.326 \\
Gemini (Nano-Banana 2)   & 0.918 & 0.971 & 0.936 & 0.927 & 0.284 & 0.239 & 0.505 & 0.433 \\
\midrule[\heavyrulewidth]
w/o UGL                  & 0.670 & 0.840 & 0.862 & 0.803 & 0.249 & 0.174 & 0.460 & 0.485 \\
Non-hierarchical         & 0.915 & 0.763 & 0.738 & 0.735 & 0.250 & 0.264 & 0.465 & 0.382 \\
w/o GEE                  & 0.609 & 0.746 & 0.822 & 0.730 & 0.218 & 0.204 & 0.518 & 0.406 \\
w/o Reinforcement        & 0.913 & 0.907 & 0.966 & 0.872 & 0.215 & 0.199 & 0.527 & 0.469 \\
\textbf{Ours} & \textbf{0.966} & \textbf{0.988} & \textbf{0.975} & \textbf{0.941} & \textbf{0.207} & \textbf{0.145} & \textbf{0.560} & \textbf{0.535} \\
\bottomrule
\end{tabular}}
\end{minipage}\hfill
\begin{minipage}[t]{0.355\textwidth}
\vspace{0pt}
\centering
\includegraphics[width=\linewidth]{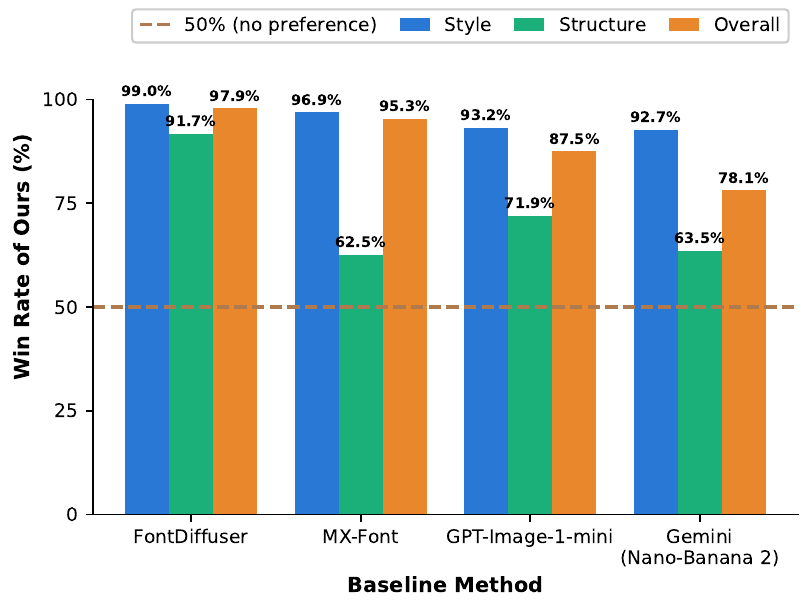}
\end{minipage}

\begin{minipage}[t]{0.62\textwidth}
\makeatletter\def\@captype{table}\makeatother
\caption{\textbf{Quantitative comparison and ablation.}
Our method most closely matches the ground truth according to LPIPS and SSIM, and achieves the highest VLM-as-a-judge scores on both font and non-font styles, across structures near to and far from the training domain. The ablations shown below degrade geometric fidelity, style fidelity, or both, with the largest performance drops occurring out of domain.}
\label{tab:quantitative}
\end{minipage}\hfill
\begin{minipage}[t]{0.345\textwidth}
\makeatletter\def\@captype{figure}\makeatother
\caption{\textbf{Human preference study.} Win rate of our method against each baseline. Our method dominates in both structure and style across all comparisons and consistently outperforms the baselines in overall preference.}
\label{fig:user_study}
\end{minipage}
\end{figure*}

\label{subsec:exp_quan_eval}

We evaluate whether our method follows the structural sketch while realizing the target style, and how well both hold as the target geometry moves away from the style's source domain.
On the in-domain tier, where paired ground truth is available, we report LPIPS for perceptual similarity to the ground truth and SSIM for agreement in local image structure.
For the near- and out-of-domain tiers, where paired ground truth is unavailable, we use
VLM-as-a-judge~\citep{pmlr-v235-chen24h}. The judge separately scores structural adherence
to the sketch and stylistic fidelity to the references; we report their product, normalized by
the score of real examples from the corresponding style
(judge setup and prompt in Appendix~\ref{sec:supp_vlm}).

\paragraph*{Comparison and ablation.}
Table~\ref{tab:quantitative} reports judge scores by style family
(\emph{Fonts}, \emph{Non-Fonts}) and by distance from the training domain
(\emph{Near}, \emph{Out}), followed by LPIPS and SSIM. Our full method achieves the best
results across all reported measures. In the upper block, the font generators closely follow the sketch but lack style fidelity,
whereas the general-purpose models better capture the target style but drift from the input
structure, as reflected in their lower \emph{Out} scores. The lower block ablates our method. \emph{w/o UGL} removes the UGL and trains the Style LoRA directly on the base model. \emph{Non-hierarchical} trains the Style LoRA without the UGL and adds the UGL only at inference. \emph{w/o GEE} trains the UGL, and applies geometry reinforcement, on original glyphs only, without GEE. \emph{w/o Reinforcement} uses the GEE-trained UGL but omits geometry reinforcement during style training. Their performance
shows that the importance of our design choices becomes more pronounced as the target geometry
moves farther from the training domain. In particular, removing geometry reinforcement lowers
the \emph{Near} score from $0.975$ to $0.966$, but causes a substantially larger drop in the
\emph{Out} score, from $0.941$ to $0.872$. This suggests that preserving geometric adherence
during style adaptation is particularly important for unfamiliar forms.

\paragraph*{User study.}
We complement these measures with user preferences collected from 32 participants, who compare
our result against each baseline in terms of style, structure, and overall preference
(Appendix~\ref{sec:supp_user_study}).
Fig.~\ref{fig:user_study} shows the same trend as Table~\ref{tab:quantitative}: our results are preferred over
every baseline on all three criteria, with the largest margins in style and overall preference.

\section{Conclusions, Limitations and Future Work}
\label{sec:conclusion}

We introduced Open-Form Geometric Style Transfer, where a geometric style
learned from one family of shapes is applied to structures well beyond that
family. We showed that learning geometric style requires separating how a shape
is realized from the particular geometry on which that style is observed.
However, \emph{disentanglement is necessary but not sufficient}. To transfer a
geometric language to unseen forms, the geometry component must itself be
universal.

These findings suggest an asymmetric view of geometry and style. Geometry
should be learned broadly, across a large and diverse space of structures, and
can be learned once as a reusable capability. Style, in contrast, can be learned
from a much narrower collection of examples and specialized independently.
Our Universal Geometry LoRA and Style LoRA instantiate this separation, while
GEE provides the geometric diversity needed to make the former universal.

As with many fine-tuned generative models, some challenges remain. Complex
structures can expose a tradeoff between geometric and stylistic fidelity, while
strong or recurring stylistic elements can occasionally interfere with the target
structure or become overemphasized. Improving this separation, particularly for
increasingly complex geometries and styles, is an interesting direction for
future work.
We believe this principle may extend beyond geometric style transfer, enabling
learned visual languages to be applied outside the domains in which they were
observed.

{
    \small
    \bibliographystyle{ieeenat_fullname}
    \bibliography{ref1}
}

\clearpage
\appendix
\makeatletter\providecommand{\setcaptypefigure}{\def\@captype{figure}}\makeatother %
\begin{figure*}[!b]
\providecommand{\fcdir}{figures/assets/appendix_further_comparisons}

    \centering
    \setlength{\tabcolsep}{1pt}
    \ifdefined\fcimw\else\newlength{\fcimw}\fi
    \setlength{\fcimw}{0.074\linewidth}
    \providecommand{\fcim}{}\renewcommand{\fcim}[1]{\includegraphics[width=\fcimw]{\fcdir/#1}}
    \providecommand{\fclab}{}\renewcommand{\fclab}[1]{\makebox[0pt][c]{#1}}
    \ifdefined\fcrefw\else\newlength{\fcrefw}\fi
    \setlength{\fcrefw}{0.85\fcimw}%
    \providecommand{\fcrefbox}{}\renewcommand{\fcrefbox}[1]{%
        \raisebox{\dimexpr(\fcimw-\fcrefw-2.8pt)/2}{%
        \tikz\node[draw=black!45, dashed, line width=0.4pt, inner xsep=3pt, inner ysep=1pt]{%
        \includegraphics[width=\fcrefw]{\fcdir/#1_ref_1.jpg}\hspace{1pt}%
        \includegraphics[width=\fcrefw]{\fcdir/#1_ref_2.jpg}\hspace{1pt}%
        \includegraphics[width=\fcrefw]{\fcdir/#1_ref_3.jpg}};}}
    \providecommand{\fcskf}{}\renewcommand{\fcskf}[1]{\fcim{#1}}
    \providecommand{\fcrow}{}\renewcommand{\fcrow}[1]{\fcskf{#1_sk.jpg} & \fcrefbox{#1} & \fcim{#1_res.jpg} & \fcim{#1_ours.jpg}}
    \providecommand{\fcblock}{}\renewcommand{\fcblock}[5]{%
        \begin{tabular}[t]{@{}c@{\hspace{2pt}}c@{\hspace{2pt}}c@{\hspace{2pt}}c@{}}
            \fcrow{#1_#2} \\[3pt]
            \fcrow{#1_#3} \\[3pt]
            \fcrow{#1_#4} \\[-1pt]
            \fclab{Sketch} & \fclab{Style Refs} & \fclab{#5} & \fclab{\bfseries\pdfliteral direct{2 Tr 0.25 w}Ours\pdfliteral direct{0 Tr 0 w}}
        \end{tabular}}
    {\fontsize{6pt}{7.5pt}\selectfont\mdseries\comfortaa
    \begin{tabular}{@{}c@{\hspace{8pt}{\color{black!40}\vrule width 0.3pt}\hspace{8pt}}c@{}}
        \fcblock{blora}{colortube}{candy}{funkyform}{B-LoRA}
        &
        \fcblock{ziplora}{leesan}{athens}{harryp}{ZipLoRA}
    \end{tabular}
    }
    \caption{\textbf{Further comparisons.} Each row shows the structural sketch, three examples of the style (framed), the result of the compared method, and ours. Left: B-LoRA~\cite{blora} transfers palette and texture but not the structure. Right: ZipLoRA~\cite{shah2023ZipLoRA}, applied to our UGL and a Style LoRA trained without it, follows the sketch but weakens the style. }
    \label{fig:supp_further}
\end{figure*}

\section{Further Comparisons}
\label{sec:supp_further}

Sec.~\ref{sec:experiments} compares against font-generation models and general-purpose image generators. Here we include additional comparisons with three recent methods that are particularly relevant to our setting. B-LoRA~\citep{blora} and ZipLoRA~\citep{shah2023ZipLoRA}, like ours, use low-rank adapters to separate what is depicted from how it is rendered, while Abstraction in Style (AiS)~\citep{lu2026abstraction} explicitly models structural abstraction as part of stylization. Fig.~\ref{fig:supp_further} shows the B-LoRA and ZipLoRA results next to ours for the same sketch and style; AiS is discussed separately below.

\paragraph*{B-LoRA.}
B-LoRA leverages LoRA to implicitly separate the style and content of a single image by jointly learning the LoRA weights of two specific SDXL blocks, which can then be used as independent components. We use the official implementation and settings, which we verified by reproducing the authors' published example. The content adapter is trained on the structural sketch and the style adapter on six training examples of the style, matching the other reference-based baselines. As Fig.~\ref{fig:supp_further} (left) shows, the style block carries only the palette and texture of the style: the geometric style is lost, and the structure of the sketch is replaced by hallucinated, unrelated content, because a thin line drawing provides little reconstruction signal.

\paragraph*{ZipLoRA.}
ZipLoRA merges independently trained style and subject LoRAs by learning per-column merger coefficients that minimize their interference. As no official implementation exists, we implement it on our own backbone, which isolates the merging strategy: the content adapter is our UGL and the style adapter is a Style LoRA trained \emph{without} the UGL, as in the \emph{Non-hierarchical} ablation (Sec.~\ref{subsec:exp_quan_eval}). Both adapters are frozen and only the coefficients are optimized, with the paper's hyperparameters. With all coefficients at one, the two adapters are simply added, which is exactly the \emph{Non-hierarchical} configuration of our ablation study; sampling uses its sketches, prompts, seeds and guidance. The learned coefficients scale down both adapters, averaging \mbox{$0.39$-$0.48$} for the UGL and \mbox{$0.24$-$0.52$} for the Style LoRA across the three styles. As Fig.~\ref{fig:supp_further} (right) shows, the result follows the sketch, but the conflict between structure and style remains, only in a milder form: style-specific details such as the pictogram tile and the wedge-shaped terminals of the display font are weakened. Training the Style LoRA with the UGL active, as we do, avoids the conflict altogether.

\textbf{Abstraction in Style (AiS).} AiS is designed specifically for styles in which abstraction extends beyond color and texture to geometric and structural deformation. AiS explicitly separates structural abstraction from visual stylization in two stages: it first transforms the target into an intermediate \emph{abstraction proxy} that reflects the structural tendencies of the reference style, and then renders this proxy into the final stylized output. Both stages are learned from a small set of visual exemplars through its Visual Analogy Transfer framework. This makes AiS particularly relevant to our setting, as it is one of the few recent methods that explicitly treats structural reinterpretation as part of style transfer.
\begin{figure}[!h]
    \centering
    \setlength{\tabcolsep}{1.5pt}
    \providecommand{\aisdir}{figures/assets/appendix_ais_comparison}
    \ifdefined\aisw\else\newlength{\aisw}\fi
    \setlength{\aisw}{\dimexpr(0.945\columnwidth-28.6pt-0.5pt)/6\relax}
    \providecommand{\aisim}{}\renewcommand{\aisim}[1]{\includegraphics[width=\aisw]{\aisdir/#1}}
    \providecommand{\aissk}{}\renewcommand{\aissk}[1]{{\setlength{\fboxsep}{0.5pt}\setlength{\fboxrule}{0.4pt}%
        \fbox{\includegraphics[width=0.5\aisw]{\aisdir/#1}}}}
    \providecommand{\aislab}{}\renewcommand{\aislab}[1]{\raisebox{\dimexpr0.5\aisw-0.5ex}{\makebox[0.055\columnwidth][r]{#1}}}
    \providecommand{\aisours}{}\renewcommand{\aisours}{\bfseries\pdfliteral direct{2 Tr 0.25 w}Ours\pdfliteral direct{0 Tr 0 w}}
    \tikzset{figsep/.style={black!40, line width=0.3pt}}
    \providecommand{\mk}{}\renewcommand{\mk}[2]{\tikz[remember picture,overlay]\coordinate(#1) at (0,#2);}
    {\fontsize{6pt}{7.5pt}\selectfont\mdseries\comfortaa
    \begin{tabular}{@{}r@{\hspace{3pt}}cc@{\hspace{4pt}{\color{black!40}\vrule width 0.3pt}\hspace{4pt}}cc@{\hspace{4pt}{\color{black!40}\vrule width 0.3pt}\hspace{4pt}}cc@{}}
        & \aissk{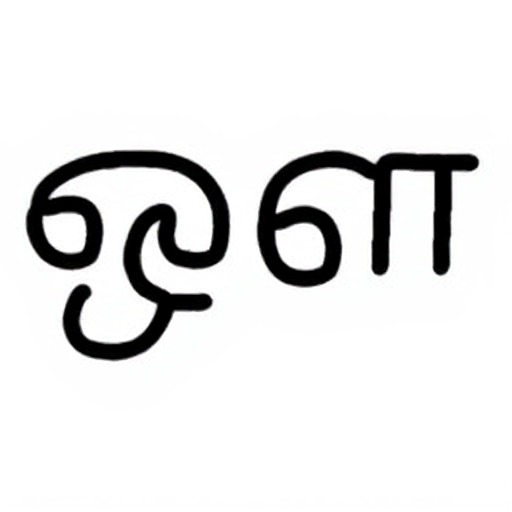} & \aissk{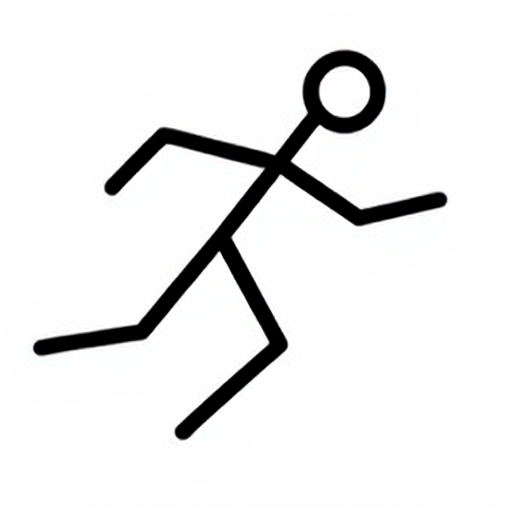} & \aissk{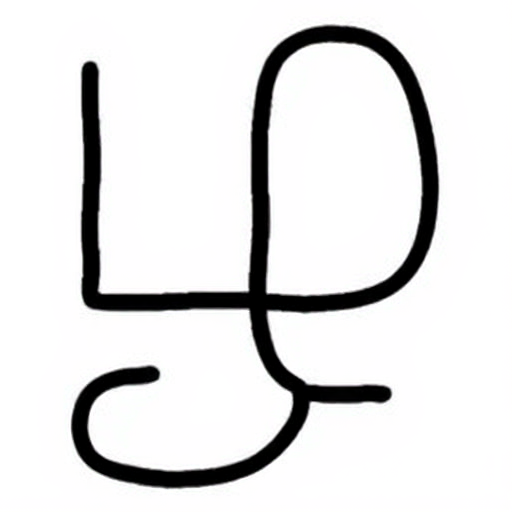} & \aissk{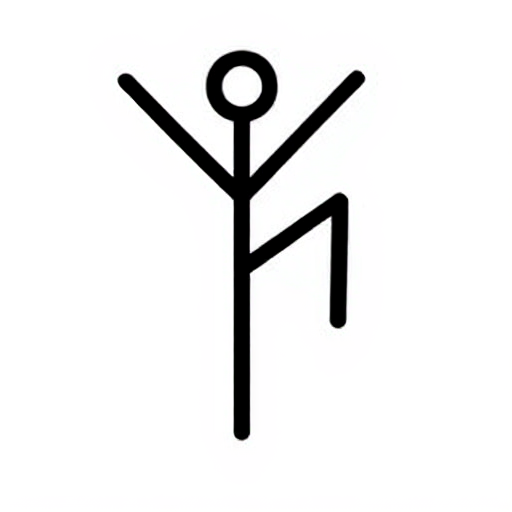} & \aissk{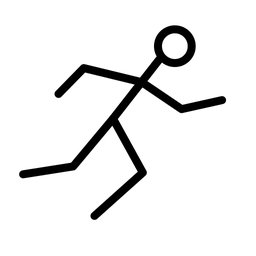} & \aissk{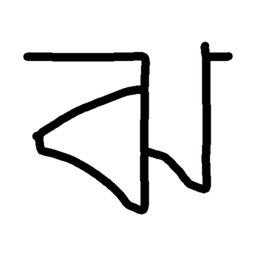} \\[4pt]
        \aislab{AiS} & \mk{hL}{0pt}\aisim{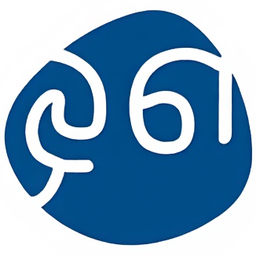} & \aisim{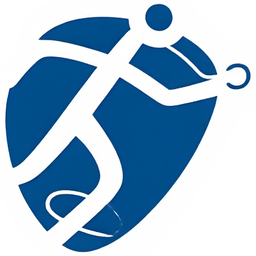} & \aisim{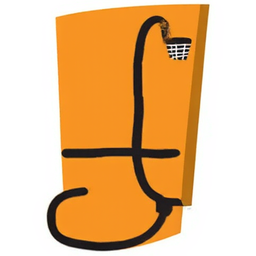} & \aisim{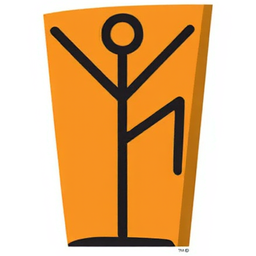} & \aisim{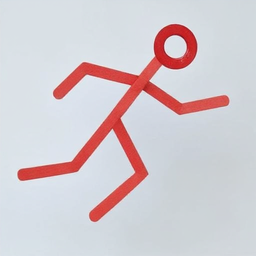} & \aisim{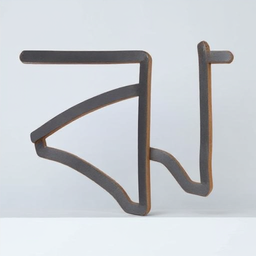}\mk{hR}{0pt} \\[8pt]
        \aislab{\aisours} & \aisim{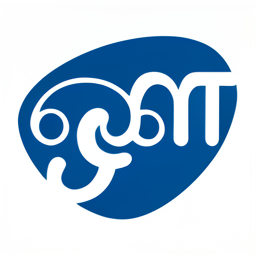} & \aisim{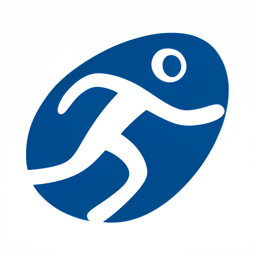} & \aisim{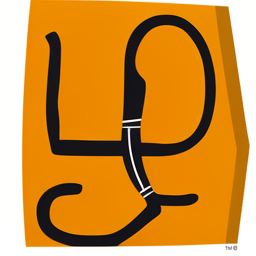} & \aisim{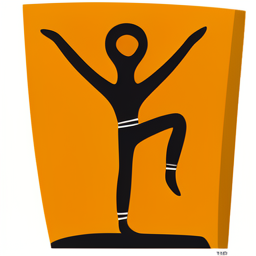} & \aisim{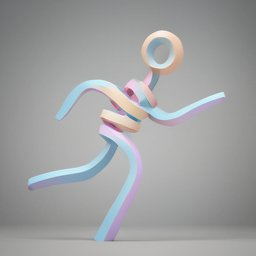} & \aisim{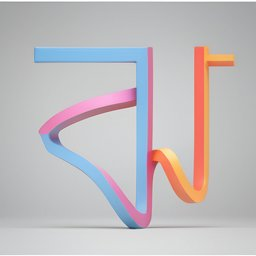}
    \end{tabular}}%
    \tikz[remember picture,overlay]\draw[figsep] ([yshift=-4.7pt]hL) -- ([yshift=-4.7pt]hR);%
    \caption{\textbf{Comparison with AiS.} For each framed sketch, the result of AiS~\citep{lu2026abstraction} (top) and ours (bottom), for two pictogram styles and a sculptural style.}
    \label{fig:supp_ais}
\end{figure}

Fig.~\ref{fig:supp_ais} compares AiS with our method on three styles: two pictogram sets and a sculptural style. We train AiS on the same style references and use the structural sketch as the abstraction target. AiS captures the visual appearance of each style, but on the pictogram styles its outputs do not reliably follow the sketch: across the scripts, strokes are rearranged, omitted, or fused, while the runner pose is only loosely preserved. On the sculptural style, whose exemplars are themselves single continuous strokes, AiS follows the sketch more closely, though it renders every glyph as one uniform tube. This reflects AiS's design: its abstraction proxy follows structural tendencies learned from the references rather than the geometry of a specific target. Our method follows each sketch closely while preserving the same style traits, including the tile deformation and stroke bands of the orange set.

\section{Dual Classifier-Free Guidance}
\label{subsec:sup_dual_guidance}

Our generator is conditioned on two inputs: a sketch image $c_I$ of the target
geometry and a text caption $c_T$ that carries the style token. Conventional text-conditioned classifier-free guidance (CFG)~\cite{ho2022classifier}
has a single scale: it amplifies the caption, while the sketch enters both
branches as an ordinary, unamplified condition, so the strength of the
geometric conditioning cannot be controlled. We therefore adopt the
compositional guidance of InstructPix2Pix~\cite{instructpix2pix}, which
separates the two conditions into terms with independent scales.

\paragraph*{Null conditions.}
CFG requires the network to model the unconditional branches, so during
training each sample is assigned to one of the four conditioning branches
listed in Table~\ref{tab:cfg_branches}: most samples keep both conditions,
one in five drops the style caption, and a small share drops the sketch or
both. The image null $\varnothing_I$ is the complete absence of reference
tokens, i.e.\ the model's native text-to-image mode, rather than a blank or
noise image, and the text null $\varnothing_T$ is the empty prompt.

\begin{table}[!h]
\centering
\footnotesize
\begin{tabular*}{\linewidth}{@{}l@{\extracolsep{\fill}}ccc@{}}
\toprule
Branch & \% of samples & Geometry $c_I$ & Style $c_T$ \\
\midrule
Fully cond.    & 70\% & $c_I$          & $c_T$          \\
Geometry cond. & 20\% & $c_I$          & $\varnothing_T$ \\
Style cond.    & 5\%  & $\varnothing_I$ & $c_T$          \\
Uncond.        & 5\%  & $\varnothing_I$ & $\varnothing_T$ \\
\bottomrule
\end{tabular*}
\caption{\textbf{Conditioning branches during training} and their sampling frequencies.}
\label{tab:cfg_branches}
\end{table}

\paragraph*{Guidance.}
Let $v_\theta(z_t, c_I, c_T)$ denote the predicted velocity at noise level $t$.
Each sampling step evaluates the unconditional, geometry-conditioned, and
fully conditioned branches of Table~\ref{tab:cfg_branches} (the style-only
branch is not used at sampling) and combines them as
\begin{equation}
\begin{aligned}
\tilde v_\theta(z_t, c_I, c_T) \;=\;& v_\theta(z_t, \varnothing_I, \varnothing_T) \\
&+ s_I \big[ v_\theta(z_t, c_I, \varnothing_T) - v_\theta(z_t, \varnothing_I, \varnothing_T) \big] \\
&+ s_T \big[ v_\theta(z_t, c_I, c_T) - v_\theta(z_t, c_I, \varnothing_T) \big].
\end{aligned}
\label{eq:dual_cfg}
\end{equation}
\noindent
The first bracket moves the prediction toward the sketch and is scaled by the
geometry scale $s_I$; the second adds the contribution of the style token on
top of the sketch and is scaled by the style scale $s_T$. With $s_I = s_T$, the two brackets merge into one and the rule becomes
single-scale CFG on the joint condition $(c_I, c_T)$. Values of $s_I$ above
$1$ amplify the geometric conditioning independently of the style scale. In practice we use $s_I \in [4, 6]$ and $s_T \in [2, 3]$
(Sec.~\ref{sec:experiments}).

\input{figures/appendix_guidance_grid}          %

\begin{figure}[t]
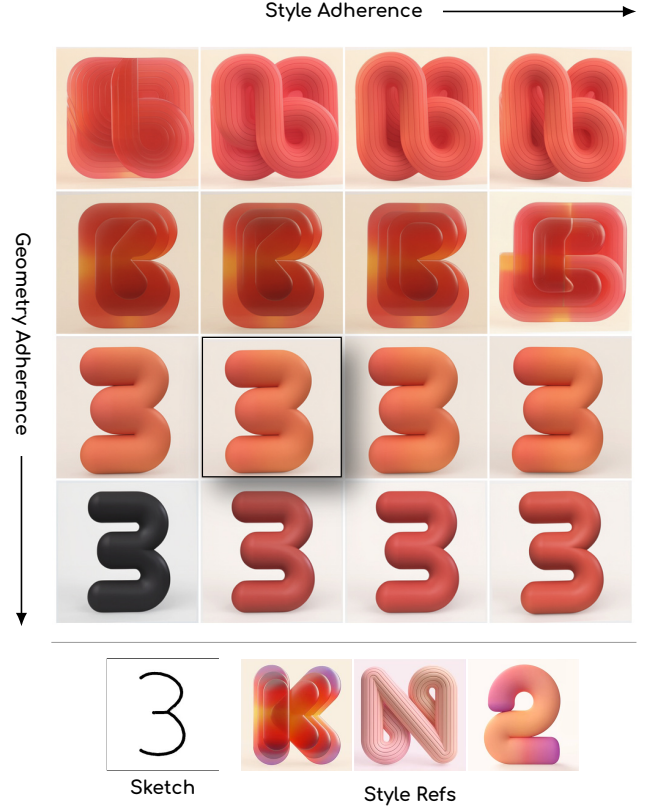

\centering
\usebox{\guidbox}
\caption{The effect of the two guidance scales of Eq.~\ref{eq:dual_cfg}; $s_T$ grows left to right, $s_I$ top to bottom. The framed cell is our default operating point. The structural sketch and three reference examples of the style are shown below the grid.}
\label{fig:guidance_grid}
\end{figure}
\noindent
\paragraph*{Reading the two levers.}
Fig.~\ref{fig:guidance_grid} jointly sweeps the two guidance scales for a
held-out sketch of a ``3''.
The grid makes the roles of the two guidance signals and their characteristic
failure modes clear.

With no sketch guidance ($s_I=0$, top row), the sketch is effectively ignored: increasing $s_T$ strengthens the style but does not recover the geometry of a ``3''. At $s_I=1$, a coherent letterform emerges, but with incorrect geometry, and strong style guidance can further distort its shape.

At the other extreme, strong sketch guidance ($s_I=8$, bottom row) reliably preserves the geometry but weakens the appearance signal. Increasing $s_T$ restores the red tubular style, though gradients and translucency remain muted. The intermediate regime ($s_I=6$, third row) provides the best balance, recovering the correct ``3'' while preserving the inflated volume and warm gradient of the exemplars. We therefore choose our operating point in this regime.

\section{Geometry Enrichment Engine: Details}
\label{sec:supp-gee}

This appendix details the construction of the geometry-enriched training corpus
used to train UGL (Sec.~\ref{sec:gee}).

\paragraph*{Generating unfamiliar geometries.}
We begin with $80$ base glyphs (Latin letters, Latin Extended letters, and digits) shared across $16$ fonts and treat them
as raw geometric material. We expand this initial vocabulary using two simple
structural operations: \emph{component removal}, which removes a structural
part of a glyph, and \emph{merging}, which combines components from two glyphs.
The operations are applied iteratively, allowing newly generated shapes to
participate in subsequent operations. As illustrated in Fig.~\ref{fig:GEM} of Sec.~\ref{sec:gee}, the
resulting forms quickly depart from recognizable characters, producing a much
broader distribution of coherent but unfamiliar geometries.

We represent each sequence of structural operations as a \emph{geometric
recipe}. Importantly, a recipe is defined independently of a particular font
and can therefore be applied to corresponding glyphs across multiple fonts.
This produces related geometric constructions realized in different styles.
Beyond increasing geometric diversity, this correspondence is what allows us to
separate the structural information conveyed by a sketch from the style of its
target realization.

A GEE recipe is a sequence of component-removal and merging operations, with both operations applied one to three times. The output of one step is the input of the next.
Both operations share a first step, graph extraction, described next.

\paragraph*{Graph extraction.}
Editing a glyph's outline directly is fragile, since a removal or a join depends on local contour details that differ from style to style. Both operators therefore act on the glyph's underlying stroke structure, which we extract as a graph; Fig.~\ref{fig:graph_extraction} of Sec.~\ref{sec:gee} illustrates the steps. Starting from the glyph (a), we compute its medial axis transform, which gives the centerline of every stroke together with the radius of the maximal inscribed disk at each point, a local estimate of stroke thickness (b). Treating the axis pixels as a graph, we prune the short branches caused by serifs and other stylistic contour details, retaining the dominant stroke structure (c). Chains of pixels between meeting points are then collapsed into single edges, junctions become nodes and stroke ends become tips (d). Finally, each edge is classified (e). An edge between two junctions is a \emph{bar}, and an edge from a junction to a tip is a \emph{leg}. Because these roles do not depend on the exact contour of a style, an operation on these graphs can be replayed across styles.

\paragraph*{Structural correspondence.}
A recipe names the component it edits, but only within the style on which it was defined. To apply the same edit consistently across styles, we must identify the corresponding component in each style. We represent the selected component using a compact descriptor consisting of its type (bar or leg), the orientation of the chord connecting its endpoints, its midpoint normalized by the glyph's bounding box, and its arc length relative to glyph size. For each other style, we compare this descriptor with all removable components of the same type and select the component with the smallest weighted distance, provided that the distance is below a matching tolerance. If no component falls below this tolerance, the recipe is not applied to that style. A recipe is retained only if at least a minimum number of styles remain (we configured as 6).

\paragraph*{Component removal.}
Fig.~\ref{fig:component_removal} of Sec.~\ref{sec:gee} illustrates the process step by step. Given the extracted graph, we aim to remove exactly one structural component, such as a leg or bar, while keeping the rest of the outline unchanged and reconstructing the cut in the font's own terminal style (a). The component to be removed is described by its structural attributes in the graph. In each style, we use these attributes to identify the corresponding component, so that the same component is removed consistently across styles. Removing only the skeleton associated with the component's edge is not sufficient, since part of its medial axis may also belong to the junction. To avoid leaving an artificial bulge, we remove medial-axis points near the junction within a radius slightly smaller than its inscribed-disk radius (b). This creates a small gap, which we reconnect with a straight stroke that smoothly bridges the surviving branches. We remove pixels belonging to the deleted limb while preserving those shared with surviving strokes. We then fill small junction concavities using the convex hull of the surviving strokes (c). The removal is defined as a pixel mask, but the output must be vectorized. We therefore trace the removal region into a smooth contour and subtract it from the original glyph outline (d). Geometry outside the edited region is preserved exactly. The newly exposed cut boundary has no counterpart in the original glyph. To reconstruct it in a style-consistent way, we search other glyphs from the same font for a matching outline segment, after normalizing for position, scale, and rotation. We then fit this donor segment to the exposed boundary and insert it into the cut region. The result is a glyph with one structural component removed while preserving the original outline wherever possible (e).

\paragraph*{Merging.}
Fig.~\ref{fig:merge_graph} of Sec.~\ref{sec:gee} illustrates the process. The connection between the two glyphs is described through structural attributes of their graphs. These attributes identify where the guest should attach to the host and allow the corresponding connection to be found across styles. We begin with two glyphs from the same style: a \emph{guest} $g_a$ (a) and a \emph{host} $g_b$ (b). Our goal is to attach the guest to the host so that they form a single coherent glyph. The merge has two stages. First, we augment the pair, \[ \hat{g}_a=\mathrm{augment}(g_a), \quad \hat{g}_b=\mathrm{augment}(g_b), \] to generate multiple candidate configurations. The host may be mirrored, while the guest may be also rotated and scaled. These choices are randomly sampled when the recipe is created. In Fig.~\ref{fig:merge_graph}(c), for example, the host is unchanged while the guest is mirrored and rotated by $270^\circ$. We then merge the augmented pair, \[ \mathrm{merge}(\hat{g}_a,\hat{g}_b). \] On the guest, the connection is anchored at a stroke end facing the host, marked by the red dot in Fig.~\ref{fig:merge_graph}(c). On the host, we select the stroke end facing the guest; if none exists, we use the nearest skeleton point (Fig.~\ref{fig:merge_graph}(d)). The guest is then translated so that the two connection points align. We keep only merges that form a coherent single shape, with a localized overlap at the junction and no narrow gaps or small enclosed holes, as shown in Fig.~\ref{fig:merge_graph}(e). A merge recipe is retained only if it succeeds across a minimum number of styles, ensuring that the structural relation transfers consistently. Small fragments extending beyond the seam are removed to produce the final glyph, as shown in Fig.~\ref{fig:merge_graph}(f).

\paragraph*{Style-suppressed structural sketches.}
For each generated glyph, we extract a structural sketch from its medial axis.
A raw medial axis, however, can retain local characteristics of the font from
which it was extracted, allowing the conditioning sketch to inadvertently
reveal style. We therefore simplify the extracted structure by pruning small
branches associated with local stylistic details and approximating the
remaining structure with a limited set of B\'ezier curves. During training, we
additionally apply random modifications to these curves. Together, these
operations preserve the dominant structure while making residual stylistic
cues unreliable.

We now describe the style-suppressing operations in more detail.

\textbf{Pruning.} We first remove branches that are more likely to arise from outline decoration than from genuine stroke structure. Pruning is defined relative to the local stroke width rather than by an absolute length threshold. In an initial pass, branches shorter than $0.4$ stroke widths are removed. This suppresses small decorative spurs without systematically eliminating genuinely short structural components. Serifs require separate treatment. At a serif terminal, the medial axis often splits into two short branches on opposite sides of the stroke. Removing either branch independently would shorten the stroke rather than remove only its decoration. We distinguish such serif pairs from genuine bifurcations using the line segment joining the two branch tips. If this segment lies entirely inside the glyph, the branches are treated as the two prongs of a serif. If it exits the glyph, as between the upper arms of a \texttt{Y}, the branches are treated as genuine structure and retained. A detected serif pair is replaced by a single branch between the two prongs, preserving the full stroke extent while discarding the terminal shape. In our fonts, serif branches span at most approximately one stroke width, whereas genuine structural limbs typically span two to five stroke widths. We therefore merge a pair only when both branches are shorter than $1.5$ stroke widths. A final pruning pass removes branches shorter than $0.9$ stroke widths, except where such a branch independently captures a substantial part of the glyph structure.

{\clubpenalty=10000 \textbf{Representational bottleneck.} We next approximate the pruned medial axis with a limited number of cubic B\'ezier curves. This sparse representation acts as an explicit information bottleneck rather than merely a compact encoding. A limited number of smooth curves can preserve the coarse glyph geometry, including stroke paths, curvature, and connectivity, while lacking the capacity to reproduce fine stylistic irregularities inherited from the original outline. Curve fitting is performed adaptively. The axis is initially represented by a single cubic B\'ezier curve and is subdivided at the point of maximum fitting error only when needed. Capacity is therefore allocated preferentially to regions that require additional geometric detail. Across our data, a glyph is represented by approximately ten curves on average and by at most 12. If this capacity is exceeded, the least significant cues are discarded as intended. The subdivision tolerance is expressed relative to stroke thickness. Because the medial axis is defined by its distance to the glyph boundary, constraining the fitting error in this way keeps the approximation geometrically consistent with the glyph interior. The resulting representation is deliberately lossy while remaining faithful to the underlying stroke geometry.\par}

Although pruning and curve compression remove much of the stylistic information, they cannot guarantee complete invariance. A sufficiently expressive model may still exploit residual correlations between the conditioning sketch and the target style. We therefore introduce two training-time mechanisms in UGL to make such residual information unreliable.

\textbf{Perturbation during training.} During training, each conditioning stroke is randomly perturbed while the corresponding target glyph remains unchanged. The model therefore does not observe a single exact skeleton as the unique conditioning representation of a target. Large-scale structure is preserved, while small local details become unreliable as predictive features. The curve representation allows us to perturb strokes structurally rather than at the pixel level. We deform the interior of each curve while keeping its endpoints fixed, preserving junctions and overall connectivity. Perturbation magnitude is scaled by curve length and correlated along the stroke, causing the stroke to bend smoothly as a whole rather than introducing high-frequency local noise. Each interior control point is displaced by a Gaussian whose standard deviation is a fraction of the curve's chord length, capped at twice that deviation so that a curve cannot fold back on itself. We resample this fraction uniformly for every sample at every epoch, from $[0.10, 0.20]$ when training UGL and from $[0.03, 0.08]$ when training a Style LoRA (Table~\ref{tab:training_details}), so the model sees a range of sketch precision rather than one fixed noise level.

\paragraph*{Cross-style pairing.}
To further prevent the sketch from serving as a style condition, we exploit
the cross-font correspondence provided by the geometric recipes. Rather than
always pairing a target glyph with its own extracted sketch, we can condition
it on a sketch extracted from another font generated by the same recipe.
Although their precise medial axes may differ, the sketch and target share the
same underlying construction while belonging to different styles. This
cross-style pairing discourages UGL from using the conditioning sketch to infer
the target style.

Together, geometry enrichment, style suppression, and cross-style pairing
produce a broad training distribution in which structural sketches primarily
communicate geometry rather than shape identity or style. Training UGL on this
corpus encourages it to follow the supplied structure, providing the universal
geometry prior required for open-form transfer.

\definecolor{promptblue}{RGB}{79,129,189}
\definecolor{promptbg}{RGB}{244,247,251}
\definecolor{promptpurple}{RGB}{125,96,170}%
\definecolor{promptpurplebg}{RGB}{247,244,251}

\newtcolorbox{promptbox}[2][]{%
  enhanced, breakable,
  colback=promptbg, colframe=promptblue,
  coltitle=white, fonttitle=\bfseries\small,
  boxrule=0.6pt, arc=1pt, left=6pt, right=6pt, top=5pt, bottom=5pt,
  title={#2}, #1
}

\newpage%
\section{Training details}
\label{sec:supp_quantitative}
\label{subsec:training_details}

Both adapters are LoRA modules~\cite{hu2022lora} on the attention projections of the FLUX.2-klein-base-4B transformer~\cite{flux2klein2026} ($80$ linear layers); the backbone, the VAE and the Qwen3 text encoder stay frozen, the text encoder in 8-bit. UGL uses rank $192$ with $\alpha=192$; every Style LoRA uses rank $32$ with $\alpha=32$ and is trained with UGL attached as a frozen adapter, never merged. Both stages use AdamW-8bit (learning rate $10^{-4}$, constant after a linear warm-up, $\beta=(0.9, 0.999)$, weight decay $0.01$, $\epsilon=10^{-6}$), an EMA of the adapter weights with decay $0.99$, gradient checkpointing, and a flow-matching objective at $256\times256$ for both the target and the sketch condition. Adapter weights are stored in fp16. Table~\ref{tab:training_details} lists the per-stage settings.

\begin{table}[!h]
\centering
\footnotesize
\begin{tabular}{lcc}
\toprule
 & UGL (stage 1) & Style LoRA (stage 2) \\
\midrule
LoRA rank / $\alpha$          & $192$ / $192$ & $32$ / $32$ \\
Optimisation steps            & $6{,}000$ & $2{,}500$ \\
Batch $\times$ grad.\ accum.  & $16\times4$ & $4\times2$ \\
Warm-up steps                 & $400$ & $50$ \\
Learning rate                 & $10^{-4}$ & $10^{-4}$ \\
Held out                      & $5\%$ recipes & $6$ rows per style \\
UGL replay in batch           & -- & $10\%$ \\
Sketch perturbation           & $10$--$20\%$ & $3$--$8\%$ \\
Hardware                      & $1\times$ H100 & $1\times$ L40S \\
\bottomrule
\end{tabular}
\caption{Training settings of the two stages.}
\label{tab:training_details}
\end{table}

\clearpage
\twocolumn[{%
\section{Additional Results}
\label{sec:supp_additional}
\setcaptypefigure
\centering
\providecommand{\gdir}{figures/assets/appendix_sculptural_grid}
\def\gres{1.135cm}%
\def\gsk{1.135cm}%
\def\gref{0.42cm}%
\def\grefh{0.64cm}
\adjustbox{totalheight=\dimexpr\textheight-82pt\relax, max width=\textwidth}{%
\begin{tikzpicture}[x=1.53cm,y=1.17cm,%
    img/.style={inner sep=0pt, outer sep=0pt},
    sk/.style={inner sep=0pt, outer sep=0pt},%
    refframe/.style={draw=black!45, dashed, line width=0.17pt, inner sep=1.5pt, dash pattern=on 1.3pt off 0.9pt}]
\node[sk] at (-0.45,0) {\includegraphics[width=\gsk]{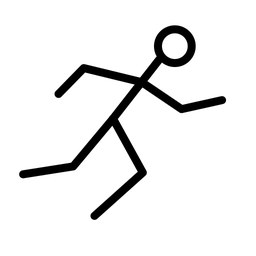}};
\node[img] at (1,0) {\includegraphics[width=\gres]{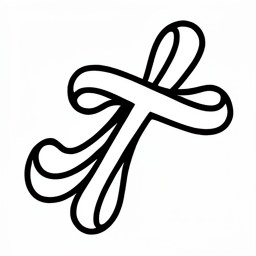}};
\node[img] at (2,0) {\includegraphics[width=\gres]{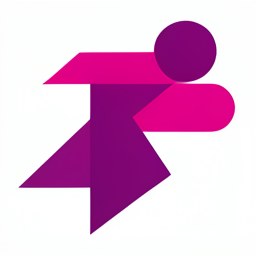}};
\node[img] at (3,0) {\includegraphics[width=\gres]{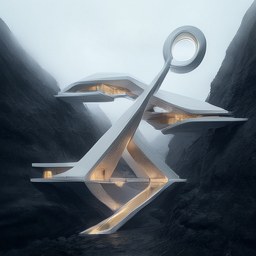}};
\node[img] at (4,0) {\includegraphics[width=\gres]{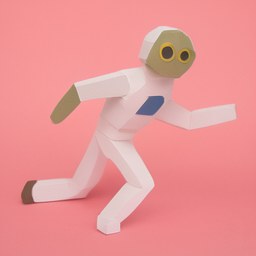}};
\node[img] at (5,0) {\includegraphics[width=\gres]{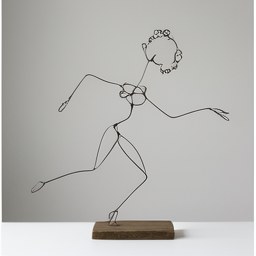}};
\node[sk] at (-0.45,-1) {\includegraphics[width=\gsk]{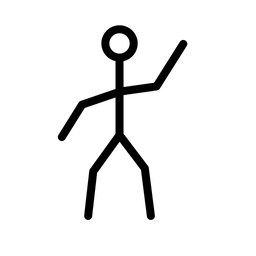}};
\node[img] at (1,-1) {\includegraphics[width=\gres]{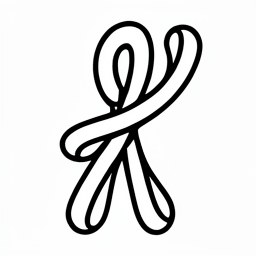}};
\node[img] at (2,-1) {\includegraphics[width=\gres]{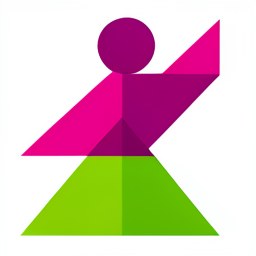}};
\node[img] at (3,-1) {\includegraphics[width=\gres]{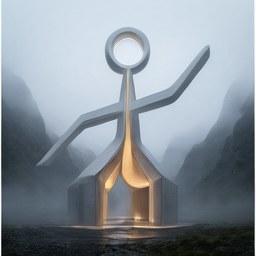}};
\node[img] at (4,-1) {\includegraphics[width=\gres]{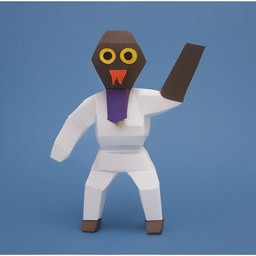}};
\node[img] at (5,-1) {\includegraphics[width=\gres]{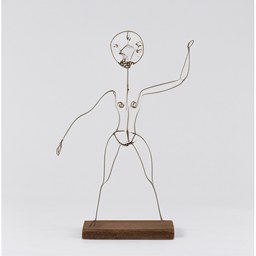}};
\node[sk] at (-0.45,-2) {\includegraphics[width=\gsk]{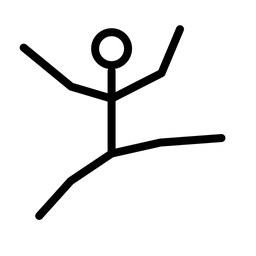}};
\node[img] at (1,-2) {\includegraphics[width=\gres]{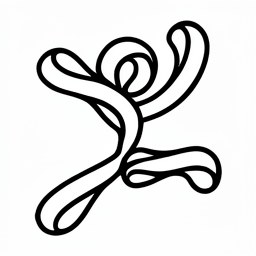}};
\node[img] at (2,-2) {\includegraphics[width=\gres]{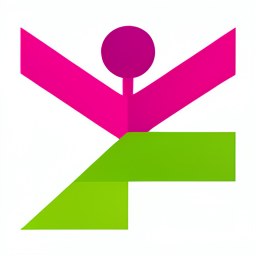}};
\node[img] at (3,-2) {\includegraphics[width=\gres]{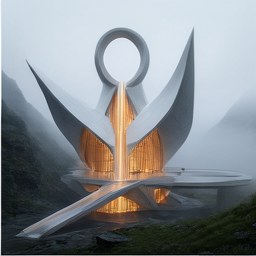}};
\node[img] at (4,-2) {\includegraphics[width=\gres]{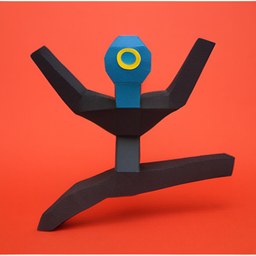}};
\node[img] at (5,-2) {\includegraphics[width=\gres]{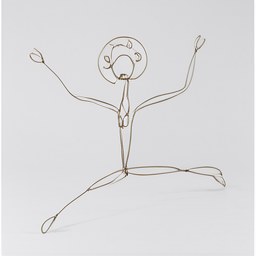}};
\node[sk] at (-0.45,-3) {\includegraphics[width=\gsk]{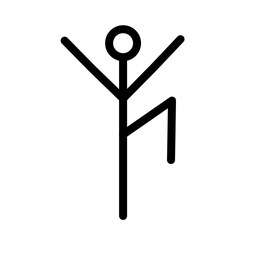}};
\node[img] at (1,-3) {\includegraphics[width=\gres]{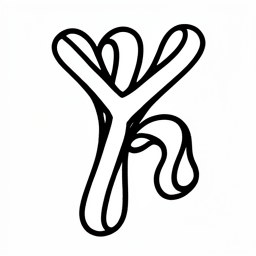}};
\node[img] at (2,-3) {\includegraphics[width=\gres]{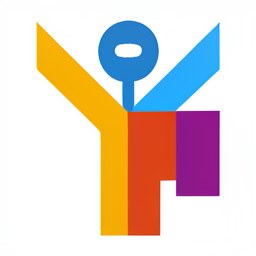}};
\node[img] at (3,-3) {\includegraphics[width=\gres]{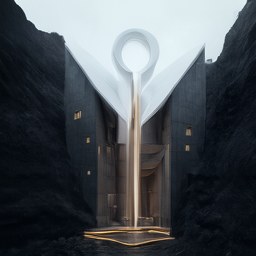}};
\node[img] at (4,-3) {\includegraphics[width=\gres]{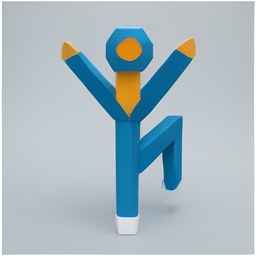}};
\node[img] at (5,-3) {\includegraphics[width=\gres]{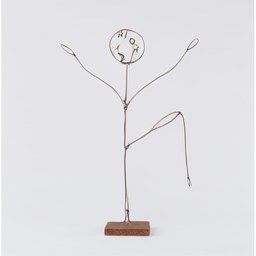}};
\node[sk] at (-0.45,-4) {\includegraphics[width=\gsk]{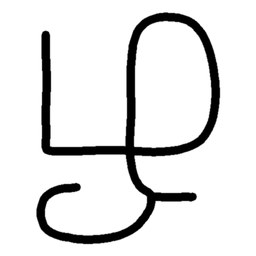}};
\node[img] at (1,-4) {\includegraphics[width=\gres]{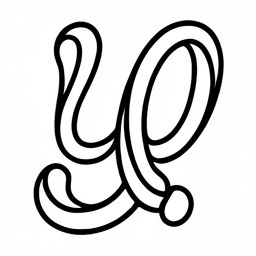}};
\node[img] at (2,-4) {\includegraphics[width=\gres]{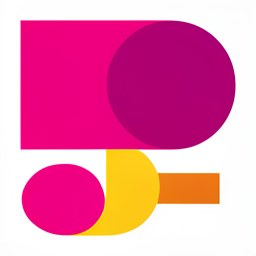}};
\node[img] at (3,-4) {\includegraphics[width=\gres]{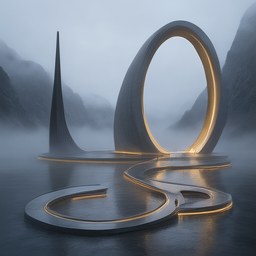}};
\node[img] at (4,-4) {\includegraphics[width=\gres]{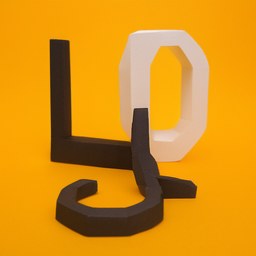}};
\node[img] at (5,-4) {\includegraphics[width=\gres]{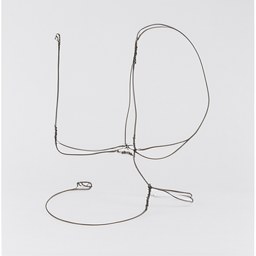}};
\node[sk] at (-0.45,-5) {\includegraphics[width=\gsk]{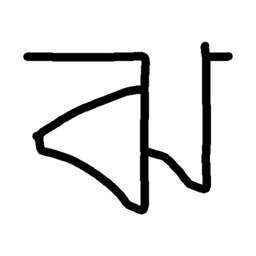}};
\node[img] at (1,-5) {\includegraphics[width=\gres]{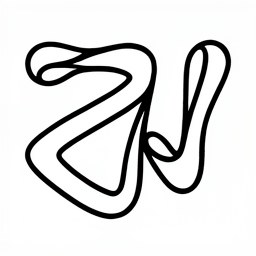}};
\node[img] at (2,-5) {\includegraphics[width=\gres]{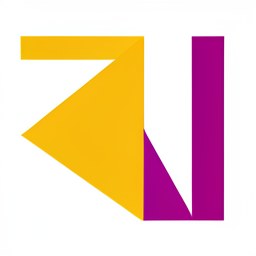}};
\node[img] at (3,-5) {\includegraphics[width=\gres]{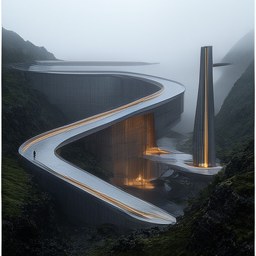}};
\node[img] at (4,-5) {\includegraphics[width=\gres]{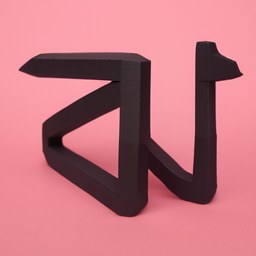}};
\node[img] at (5,-5) {\includegraphics[width=\gres]{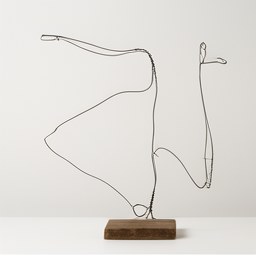}};
\node[sk] at (-0.45,-6) {\includegraphics[width=\gsk]{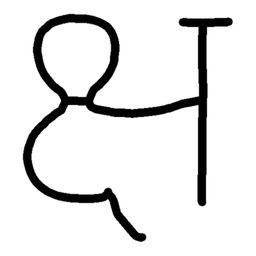}};
\node[img] at (1,-6) {\includegraphics[width=\gres]{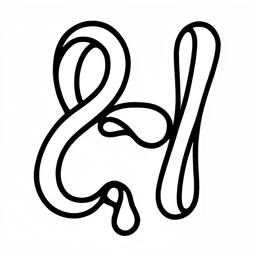}};
\node[img] at (2,-6) {\includegraphics[width=\gres]{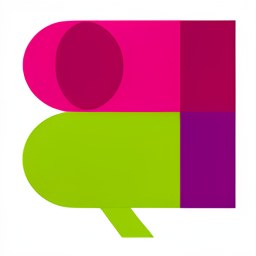}};
\node[img] at (3,-6) {\includegraphics[width=\gres]{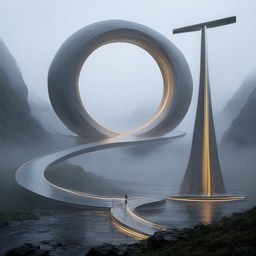}};
\node[img] at (4,-6) {\includegraphics[width=\gres]{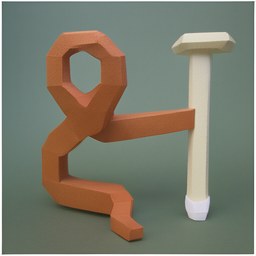}};
\node[img] at (5,-6) {\includegraphics[width=\gres]{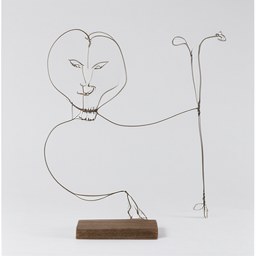}};
\node[sk] at (-0.45,-7) {\includegraphics[width=\gsk]{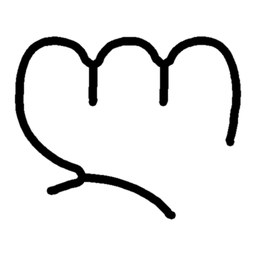}};
\node[img] at (1,-7) {\includegraphics[width=\gres]{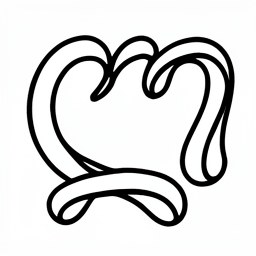}};
\node[img] at (2,-7) {\includegraphics[width=\gres]{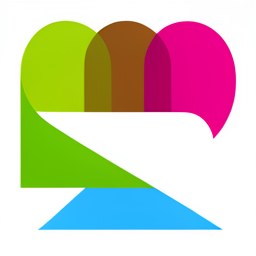}};
\node[img] at (3,-7) {\includegraphics[width=\gres]{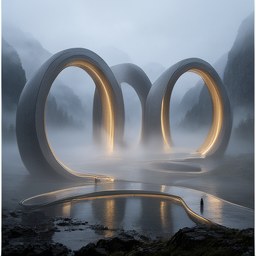}};
\node[img] at (4,-7) {\includegraphics[width=\gres]{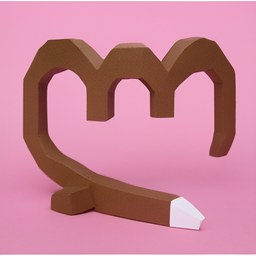}};
\node[img] at (5,-7) {\includegraphics[width=\gres]{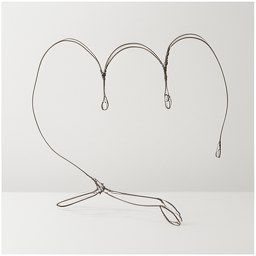}};
\node[sk] at (-0.45,-8) {\includegraphics[width=\gsk]{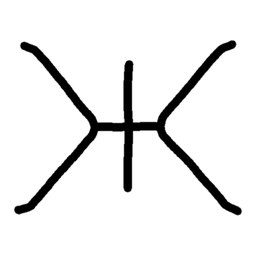}};
\node[img] at (1,-8) {\includegraphics[width=\gres]{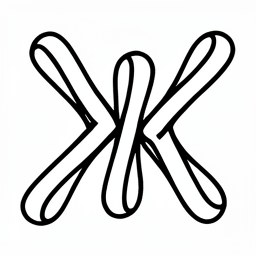}};
\node[img] at (2,-8) {\includegraphics[width=\gres]{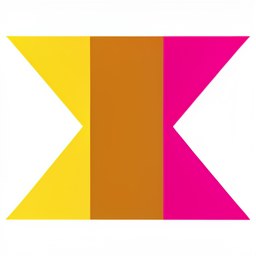}};
\node[img] at (3,-8) {\includegraphics[width=\gres]{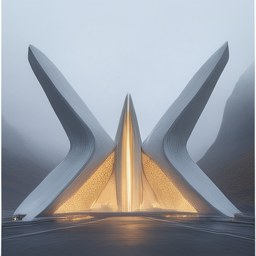}};
\node[img] at (4,-8) {\includegraphics[width=\gres]{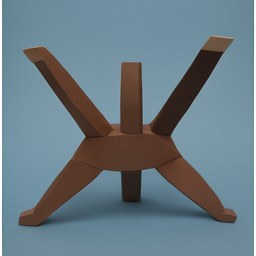}};
\node[img] at (5,-8) {\includegraphics[width=\gres]{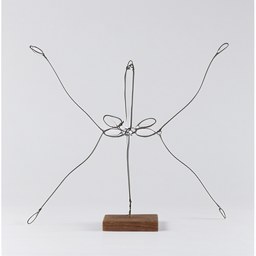}};
\node[img, xshift=-0.44cm] (r10) at (1,0.9) {\includegraphics[width=\gref,height=\grefh,keepaspectratio]{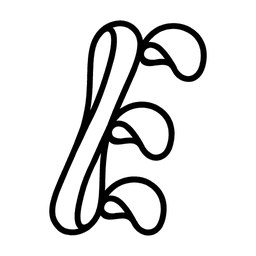}};
\node[img, xshift=0.00cm] (r11) at (1,0.9) {\includegraphics[width=\gref,height=\grefh,keepaspectratio]{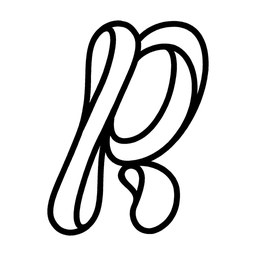}};
\node[img, xshift=0.44cm] (r12) at (1,0.9) {\includegraphics[width=\gref,height=\grefh,keepaspectratio]{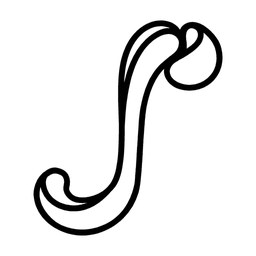}};
\node[img, xshift=-0.44cm] (r20) at (2,0.9) {\includegraphics[width=\gref,height=\grefh,keepaspectratio]{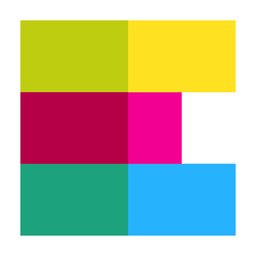}};
\node[img, xshift=0.00cm] (r21) at (2,0.9) {\includegraphics[width=\gref,height=\grefh,keepaspectratio]{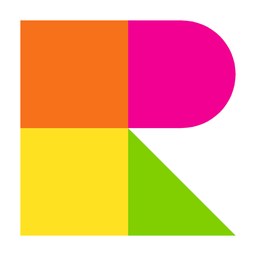}};
\node[img, xshift=0.44cm] (r22) at (2,0.9) {\includegraphics[width=\gref,height=\grefh,keepaspectratio]{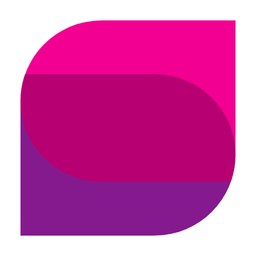}};
\node[img, xshift=-0.44cm] (r30) at (3,0.9) {\includegraphics[width=\gref,height=\grefh,keepaspectratio]{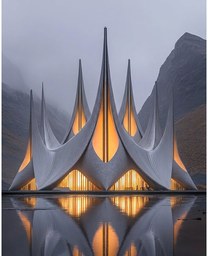}};
\node[img, xshift=0.00cm] (r31) at (3,0.9) {\includegraphics[width=\gref,height=\grefh,keepaspectratio]{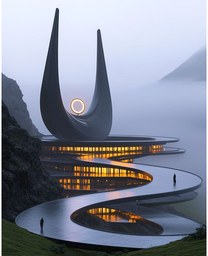}};
\node[img, xshift=0.44cm] (r32) at (3,0.9) {\includegraphics[width=\gref,height=\grefh,keepaspectratio]{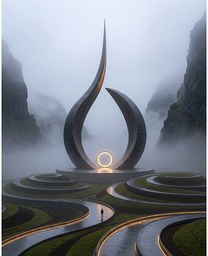}};
\node[img, xshift=-0.44cm] (r40) at (4,0.9) {\includegraphics[width=\gref,height=\grefh,keepaspectratio]{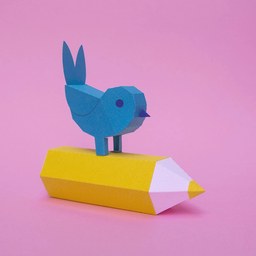}};
\node[img, xshift=0.00cm] (r41) at (4,0.9) {\includegraphics[width=\gref,height=\grefh,keepaspectratio]{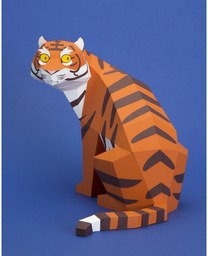}};
\node[img, xshift=0.44cm] (r42) at (4,0.9) {\includegraphics[width=\gref,height=\grefh,keepaspectratio]{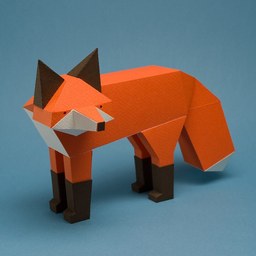}};
\node[img, xshift=-0.44cm] (r50) at (5,0.9) {\includegraphics[width=\gref,height=\grefh,keepaspectratio]{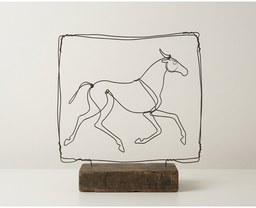}};
\node[img, xshift=0.00cm] (r51) at (5,0.9) {\includegraphics[width=\gref,height=\grefh,keepaspectratio]{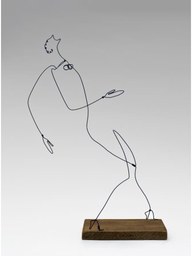}};
\node[img, xshift=0.44cm] (r52) at (5,0.9) {\includegraphics[width=\gref,height=\grefh,keepaspectratio]{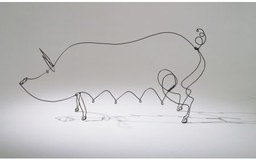}};
\foreach \y in {0,-1,-2,-3,-4,-5,-6,-7,-8} \node[img] at (0.28,\y) {\includegraphics[width=0.5cm]{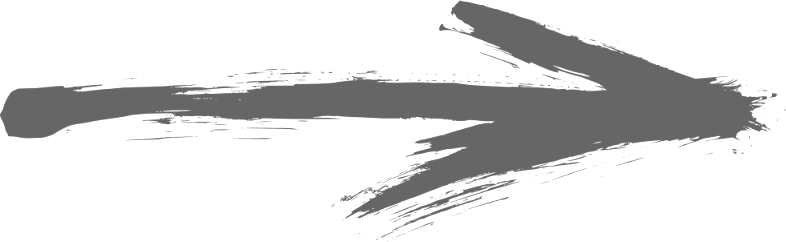}};
\node[refframe, fit=(r10) (r11) (r12)] {};
\node[refframe, fit=(r20) (r21) (r22)] {};
\node[refframe, fit=(r30) (r31) (r32)] {};
\node[refframe, fit=(r40) (r41) (r42)] {};
\node[refframe, fit=(r50) (r51) (r52)] {};
\end{tikzpicture}}
\caption{\textbf{Sculptural and decorative styles on pose and glyph sketches.} Each row is one structural sketch (left): four pose sketches (top) and five glyph sketches (bottom); each column is one style, shown by three reference examples at the top.}
\label{fig:supp_sculpt_grid_poses}\label{fig:supp_sculpt_grid}
\vspace{6pt}}]

\input{figures/appendix_additional_results}
\renewcommand{\floatpagefraction}{0.9}
\input{figures/appendix_sculptural_results}

\providecommand{\promptfont}{\fontfamily{ComicNeue-TLF}\selectfont}
\providecommand{\promptsize}{\fontsize{7.5}{8}}%

\section{VLM-as-a-Judge Evaluation}
\label{sec:supp_vlm}

We employ a VLM-as-a-Judge evaluation framework to assess generated results in the open-set regimes, where no unique paired ground truth exists: for a target structure outside the style's own shape family there is no single correct rendering to compare against, so paired metrics such as LPIPS and SSIM are undefined. The judge is asked instead whether the requested \emph{structure} is depicted and whether it is realized in the demonstrated \emph{geometric design language}.

Each judged candidate is presented as a single composite image with labeled panels: a top row of $K=6$ style references, and a bottom row holding the target structural sketch and the candidate. References are drawn only from the style's training rows, so the judge never sees the ground-truth target of the case it is grading. All panels are rendered at $256\times256$ on a common canvas, and nothing in the image or the prompt identifies the method that produced the candidate.

We score each candidate \emph{absolutely} rather than in pairs. This has two consequences we rely on: the resulting scalar does not depend on which baselines happen to be included in the comparison, and there is no left--right position bias to mitigate. To remove the remaining ordering effect, the six style references are shuffled per case with a fixed per-case permutation. The judge is instructed to return a single JSON object conforming to a strict response schema, which prevents the answer from drifting into prose and bounds the output length.

\begin{figure}[!tp]
\begin{promptbox}[unbreakable, before skip=0pt]{VLM-as-a-Judge: Evaluation Prompt (abridged)}%

\promptsize\promptfont
\setlength{\parskip}{0.5em}\setlength{\parindent}{0pt}
\textbf{System prompt}: meticulous visual judge for a geometric style transfer benchmark; sees one composite image with labeled panels; answers only in schema-conforming JSON.

\noindent\rule{\linewidth}{0.5pt}\par\vspace{1pt}

\textbf{User prompt}:

\textbf{Image layout.} Top row: 6 \textbf{Style References}, defining \textbf{how} things are drawn (strokes, curvature, corners, joins, color, texture, motifs). Bottom row: the \textbf{Target Sketch}, defining \textbf{what} must be depicted, and the \textbf{Candidate}.

\textbf{Questions}, answered in order:

\hangindent=1.5em\hangafter=1 1. \textbf{geometry\_deviations}: list every structural difference from the sketch, each tagged \emph{core} (element missing, added, moved or reconnected) or \emph{minor} (element slightly shifted or reshaped).

\hangindent=1.5em\hangafter=1 2. \textbf{geometry\_score (1--5)}: is the sketch's structure depicted? Deviations the references explain as style are allowed; style is not judged here. \textbf{5} = fully recognizable $\ldots$ \textbf{1} = a different object.

\hangindent=1.5em\hangafter=1 3. \textbf{style\_score (1--5)}: does the candidate belong to the references' visual family? Depicting the same objects is not rewarded; geometry is not judged here. \textbf{5} = could pass as a reference $\ldots$ \textbf{3} = several traits missing $\ldots$ \textbf{1} = no resemblance.

\hangindent=1.5em\hangafter=1 4. \textbf{Flags}: \textbf{source\_leakage}, the candidate reproduces a shown reference instead of the sketch; \textbf{sketch\_copy}, the sketch passed through with little or no style.

\textbf{Output}: one JSON object,
\{"geometry\_deviations": [\{"what", "severity"\}], "geometry\_score", "style\_score", "source\_leakage", "sketch\_copy"\}.
\end{promptbox}

\begin{promptbox}[unbreakable, after skip=0pt]{VLM-as-a-Judge: Example Query}
\promptsize\promptfont
\setlength{\parskip}{0.5em}\setlength{\parindent}{0pt}
\textbf{Image input}: one composite, sent alongside the prompt above. Here the
style is a pictogram set and the requested structure is a cyclist: none of the
six references depicts cycling, so the judge must separate the design language
--- white figure, dark badge, uniform stroke --- from the particular sports
that carry it.

\noindent\includegraphics[width=\linewidth]{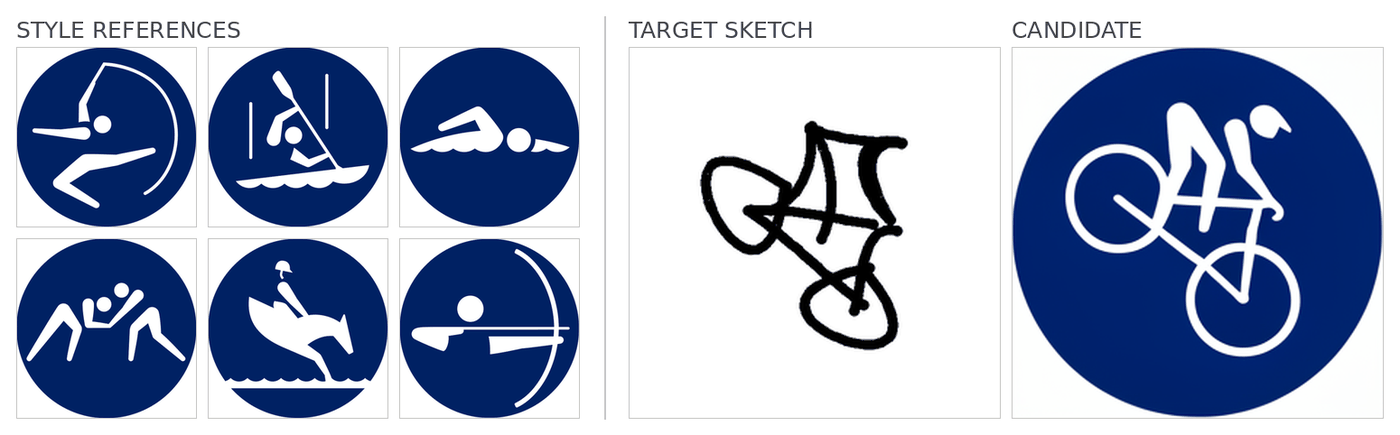}\par

\nointerlineskip\vspace{3pt}\hrule height 0.5pt\relax\vspace{2pt}%

\textbf{Response}:

\{"geometry\_deviations": [\{"what": "added head", "severity": "core"\},
\{"what": "modified frame", "severity": "minor"\}], "geometry\_score": 4,
"style\_score": 4, "source\_leakage": false, "sketch\_copy": false\}
\end{promptbox}
\end{figure}

The response to the example query illustrates the two axes being scored independently. Geometry is
$4$ rather than $5$ because the judge locates two nameable departures from the
sketch before scoring it: the rider is given the head the sketch leaves
implicit, and the bicycle frame is redrawn cleaner than the tangle of strokes
it was traced from. Style is $4$ because the candidate carries the references'
badge, weight and figure convention onto a sport none of them shows, while
stopping short of being mistaken for one of them. Both flags are false: the
candidate reproduces no shown reference, and it is not the sketch passed
through unstyled.

\subsection{Judge Configuration}
We use the same judge configuration for the entire evaluation: all reported numbers come from gemini-3.1-pro-preview with a structured response schema and medium media resolution. Each candidate is judged in a single pass.
Scores are aggregated hierarchically, matching the rest of our protocol: samples within a case, cases within a style, then styles weighted equally.

\section{User Study}
\label{sec:supp_user_study}

\subsection{User Study Setup}
We complement the VLM-as-a-Judge evaluation with a human preference study. Unlike the judge, which scores each candidate \emph{absolutely}, the study is \emph{pairwise}: participants see two results for the same input and indicate which one is better. Relative judgments are easier for non-expert participants to make consistently than absolute ratings, and they directly answer the question of interest, namely whether our results are preferred over those of each baseline.

Each question is presented as a single composite image with labeled panels: a top row with the style references and the structural sketch, and a bottom row with two results, A and B. One result is ours and the other is produced by a baseline, both generated for the same style and the same sketch. All panels, including the two results, are rendered at the same size on a common canvas, so that the native output resolution of a method is not revealed, and nothing in the image or the text identifies the method behind a result.

The study contains $12$ questions: three for each of the four baselines (MX-Font, FontDiffuser, Gemini, and GPT-Image), each on a different style, covering black-and-white fonts, decorative fonts, and pictogram styles. The sketches span held-out glyphs, non-Latin glyphs, human poses, and symbols. For each question, participants answer three multiple-choice items, on style, structure, and overall preference, each with the options \emph{Result A}, \emph{Result B}, and \emph{About the same}.

To control for ordering and position effects, we prepare four editions of the questionnaire. Each edition shuffles the order of the questions, never placing the same baseline in two consecutive questions, and randomizes which side shows our result. Within every edition our result appears as A in exactly half of the questions, and the side assignment is mirrored between editions, so that any preference for a side cancels out. Participants reach the study through a single link that assigns them to one of the editions at random. Responses are anonymous, and each participant answers one edition.

\begin{promptbox}[colframe=promptpurple, colback=promptpurplebg]{User Study: Instructions}
\promptsize\promptfont
\setlength{\parskip}{0.5em}\setlength{\parindent}{0pt}
Thank you for participating!

This study evaluates \textbf{geometric style transfer}: the task of applying the design language shared by a few style references to a new, unseen structure given by an input sketch. The sketch specifies the target structure, whereas the references specify the style in which that structure should be realized.

Each of the 12 questions shows:

{\setlength{\parskip}{0pt}\leftskip=1.5em\relax%
$\bullet$ the style references (target style)\par
$\bullet$ the structural sketch (target structure)\par
$\bullet$ two results, A and B, generated by different methods\par}

For each question, choose which result (A, B, or about the same):

{\setlength{\parskip}{0pt}\leftskip=1.5em\relax%
$\bullet$ \textbf{Style} -- better matches the style of the references.\par
$\bullet$ \textbf{Structure} -- better depicts the structure of the sketch.\par
$\bullet$ \textbf{Overall} -- you prefer overall.\par}

Look closely at all the style references: a style is often defined by distinctive motifs that appear in only some of them.
\end{promptbox}

\begin{promptbox}[colframe=promptpurple, colback=promptpurplebg]{User Study: Example Question}
\promptsize\promptfont
\setlength{\parskip}{0.5em}\setlength{\parindent}{0pt}
\textbf{Image shown}: sport pictograms as the style and a treble clef as the structure (out of domain). Result~A is ours and Result~B is Gemini; other editions swap the sides.

\begin{center}
\includegraphics[width=\linewidth]{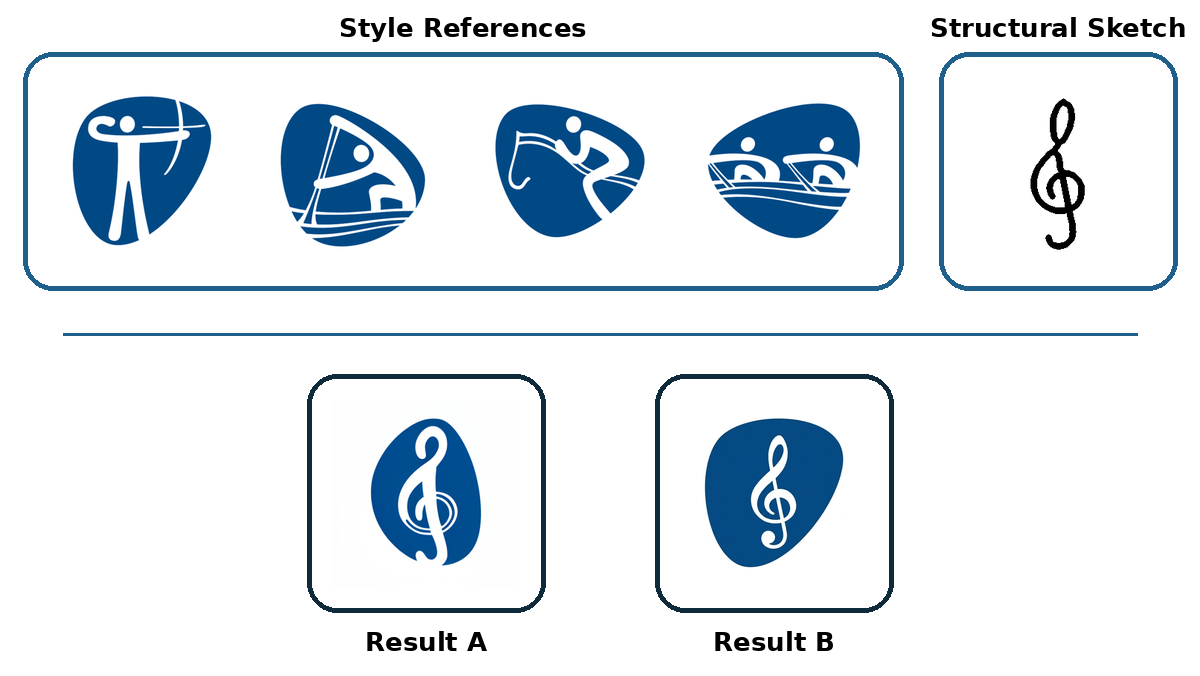}
\end{center}

\noindent\rule{\linewidth}{0.5pt}\par\vspace{1pt}

\textbf{Items} (Result A / Result B / About the same): \textbf{Style}, closer to the references? \textbf{Structure}, closer to the sketch? \textbf{Overall}, preferred?
\end{promptbox}

The example illustrates why the two axes are asked separately. Both results depict a recognizable treble clef, so the structure item mainly asks whether the particular structure of the sketch is preserved, such as its proportions, the spiral around the stem, and the hooked tail. The style item asks something different: whether the result carries the distinctive traits of the references, such as the rounded, asymmetric blue tile, the bold white strokes with rounded ends, and the thin parallel lines that accompany the figures, rather than merely sharing their colors.

Before the first question, participants are shown one worked example in the same layout, with the intended answers marked and a one-line explanation of each. Its two results are not method outputs: one is a letter of a reference font and the other is the same letter in a plain serif font, so the intended answers are unambiguous and do not bias participants toward any method.

\subsection{Aggregation}
For each baseline and each of the three items, we report the percentage of votes that prefer our result, prefer the baseline's result, or rate the two as about the same, pooled over participants, editions, and the three questions of that baseline. Significance is assessed with a two-sided exact binomial test against $50\%$ on the decisive votes, i.e., after discarding \emph{About the same} answers. In total, 32 participants completed the study.

\end{document}